\documentclass[11pt]{article}

\usepackage[preprint]{acl}
\usepackage{amsthm}

\usepackage{times}
\usepackage{latexsym}

\usepackage[T1]{fontenc}

\usepackage[utf8]{inputenc}

\usepackage{microtype}

\usepackage{inconsolata}

\usepackage{graphicx}
\usepackage{amsmath}
\usepackage{algorithm}
\usepackage{algpseudocode} 

\usepackage{booktabs}
\usepackage{multirow}
\usepackage{threeparttable}
\usepackage{adjustbox}

\usepackage{graphicx}
\usepackage{subcaption}
\usepackage[most]{tcolorbox}
\usepackage{enumitem}
\usepackage{cuted}

\newtcolorbox{promptbox}[2][]{%
  enhanced,
  breakable,
  width=\linewidth,
  colback=white,
  colframe=black,
  boxrule=0.8pt,
  arc=0pt, outer arc=0pt,
  left=6pt, right=6pt, top=6pt, bottom=6pt,
  title={#2},
  colbacktitle=gray!15,
  coltitle=black,
  fonttitle=\bfseries,
  boxed title style={
    colframe=black,
    boxrule=0.8pt,
    arc=0pt, outer arc=0pt,
    left=6pt, right=6pt, top=2pt, bottom=2pt,
  },
  attach boxed title to top left={xshift=0mm,yshift=-1mm},
  borderline south={0.6pt}{0pt}{black},
  before skip=8pt,
  after skip=8pt,
  #1
}

\newtheorem{theorem}{Theorem}[section]
\newtheorem{lemma}{Lemma}[section]

\usepackage{amsmath,amssymb}
\newtheorem{remark}{Remark} 
\newtheorem{assumption}{Assumption}

\author{
\makebox[\textwidth][c]{%
\begin{minipage}{0.95\textwidth}\centering
Xiaoxu Ma$^{1}$, Xiangbo Zhang$^{1}$, Zhenyu Weng$^{2}$\thanks{Corresponding author.}\\[0.35em]
$^{1}$Georgia Institute of Technology \quad
$^{2}$South China University of Technology\\[0.55em]
{\ttfamily\small
\begin{minipage}{0.85\textwidth}\centering
\url{xma394@gatech.edu}\quad
\url{xiangbo.zhang@gatech.edu}\quad
\url{wzytumbler@gmail.com}
\end{minipage}}
\end{minipage}
}%
}

\begin{document}
\newcommand{\corpusname}{\textsc{PVNI}}
\title{\textbf{Stable and Explainable Personality Trait Evaluation in Large Language Models with Internal Activations}}

\maketitle

\begin{abstract}

Evaluating personality traits in Large Language Models (LLMs) is key to model interpretation, comparison, and responsible deployment. However, existing questionnaire-based evaluation methods exhibit limited stability and offer little explainability, as their results are highly sensitive to minor variations in prompt phrasing or role-play configurations. To address these limitations, we propose an internal-activation–based approach, termed Persona-Vector Neutrality Interpolation (PVNI), for stable and explainable personality trait evaluation in LLMs. PVNI extracts a persona vector associated with a target personality trait from the model’s internal activations using contrastive prompts. It then estimates the corresponding neutral score by interpolating along the persona vector as an anchor axis, enabling an interpretable comparison between the neutral prompt representation and the persona direction. We provide a theoretical analysis of the effectiveness and generalization properties of PVNI. Extensive experiments across diverse LLMs demonstrate that PVNI yields substantially more stable personality trait evaluations than existing methods, even under questionnaire and role-play variants.


\end{abstract}
\section{Introduction}

Personality testing is widely recognized as a standardized tool for describing stable individual differences to improve self-understanding, communication, and decision-making \citep{personalitytest, personalitymeasure}. Extending this perspective, recent research has explored personality testing in Large Language Models (LLMs) \citep{bodroza-dinic-bojic-2024-personality, liu-lu-he-zhang-2025-many-dimensions, jiang-zhang-cao-breazeal-roy-kabbara-2024-personallm, Tang:Yang:Abbasi:arXiv2509.07370}. These studies view LLMs as agents that exhibit stable, human-interpretable behavioral tendencies, which can be quantified into personality trait profiles. Such trait-based characterizations enable the interpretation, comparison, and alignment of model behavior in real-world deployments \citep{bhandari-naseem-datta-fay-nasim-2025-evaluating, wang-zhao-ones-he-xu-2025-evaluating}.

\begin{figure*}[t]
  \centering
  \includegraphics[width=1.0\linewidth]{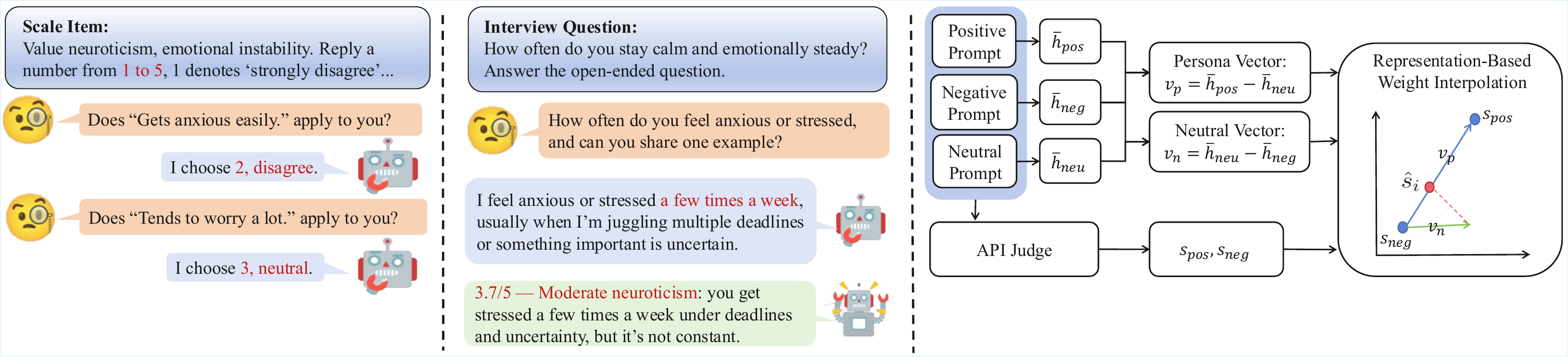}
  \caption{Comparison of self-assessment, questionnaires, and PVNI. Self-assessment and questionnaires are prompt-sensitive, while PVNI is stable and explainable with internal activations.}
  \label{fig:comparison}
  \vspace{-2mm}
\end{figure*}

Recent studies highlight the potential of personality tests for LLMs, enabling quantitative, comparable profiling of stable behavioral tendencies to support model evaluation, alignment, and controllable personalization \citep{SerapioGarcia2025PsychometricLLMPersonality, PsychometricEvaluation, zhu2025evaluatingllmalignmentpersonality}. This paradigm mainly takes two forms. First, self-report assessment \citep{han2025personalityillusionrevealingdissociation, jiang-zhang-cao-breazeal-roy-kabbara-2024-personallm} asks LLMs to directly rate Likert-style trait items, making the procedure simple, low-cost, and highly standardized for scalable comparisons. Second, open-ended questionnaires with scoring \citep{zheng-wang-hosio-xu-lee-2025-lmlpa} collect free-form responses. These responses are then mapped to trait scores via an external judge, which better reflects naturalistic behavior and captures richer, behavior-relevant trait signals under a consistent rubric.

Despite these advantages, both approaches ultimately depend on eliciting answers to researcher-designed prompts, which introduces two fundamental limitations. First, questionnaire-based measurements are often highly unstable: even minor changes in prompt framing or wording can substantially shift estimated trait scores despite no underlying change in the model \citep{zheng-wang-hosio-xu-lee-2025-lmlpa, tosato-helbling-mantilla-ramos-hegazy-tosato-lemay-rish-dumas-2025-persistent, shu-zhang-choi-dunagan-logeswaran-lee-card-jurgens-2024-unreliable}. Second, these protocols primarily quantify prompted role play through input–output behavior \citep{gupta-song-anumanchipalli-2024-self-assessment, de-wynter-wang-sokolov-gu-chen-2023-discourse-memorization}, thereby conflating performative personas with intrinsic dispositions. 
As illustrated in Figure~\ref{fig:comparison}, semantically equivalent self-report items yield different ratings for Neuroticism, while open-ended scoring produces yet another estimate. This discrepancy highlights the intrinsic instability of prompt-dependent personality measurements. Consequently, such methods offer limited evidence that the resulting trait scores reflect stable internal properties or correspond to identifiable structures within the model’s representations. Moreover, the process by which these scores are obtained remains difficult to interpret.

To overcome these limitations, we introduce an internal-activation–based approach, termed Persona-Vector Neutrality Interpolation (PVNI), for stable and explainable personality trait evaluation in LLMs. PVNI quantifies trait strength with weight interpolation in the model’s internal activation space. Specifically, it derives a persona vector for a target personality trait using contrastive prompts (positive and negative). The neutral trait score is then estimated by anchoring the neutral prompt along this persona vector and performing interpolation, yielding an interpretable comparison between the neutral prompt representation and the persona direction. We further develop a linear theory of persona vectors, showing that activation differences induced by persona prompts define directions that behave approximately linearly. Extensive experiments demonstrate that PVNI achieves substantially greater stability in personality trait evaluation than existing methods under the Big Five (OCEAN) evaluation framework. 

Our contributions are summarized as follows:
\begin{itemize}
\item We introduce Persona-Vector Neutrality Interpolation (PVNI), an internal-activation-based weight interpolation method for stable and explainable personality trait evaluation in LLMs.
\item We provide a linear theory of persona vectors establishing the effectiveness and generalization properties of PVNI.
\item Extensive experiments demonstrate that PVNI achieves more stable personality trait evaluations across diverse LLMs, even under questionnaire and role-play variants. 
\end{itemize}

\section{Related Works}
\vspace{-1mm}

\paragraph{Personality Trait Evaluation.} In practice, most work adopts the Big Five (OCEAN) framework \citep{Li:Liu:Liu:Zhou:Diab:Sap:ACL25:BIG5CHAT} as the dominant and widely used personality model for such evaluations. Current Big Five (OCEAN) assessment for LLMs largely falls into two paradigms. The first is self-report assessment, where the model is treated as a survey respondent: standardized instruments such as IPIP-NEO are administered and responses are scored using established rubrics \citep{SerapioGarcia2025PsychometricLLMPersonality, Jiang2023MPI}. The second paradigm uses open-ended elicitation \citep{zheng-wang-hosio-xu-lee-2025-lmlpa, Sandhan:Cheng:Sandhan:Murawaki:FindingsEMNLP25:CAPE}, prompting the model to produce free-form answers and then applying an automated judge to map each response to a continuous trait-intensity score. Both paradigms are highly prompt- and role-play-sensitive: minor wording or framing changes can substantially shift the measured result. In contrast, our PVNI approach estimates trait score and directions from internal activations, producing a trait signal that is substantially more robust to semantically equivalent prompt rewrites.

\paragraph{Persona Vector.}
Recent work represents LLM persona with persona vectors, interpretable directions that amplify or attenuate a target trait. \citet{sun-etal-2025-personality} define Personality Vectors in internal representation space as differences between persona-finetuned and base models, enabling continuous control and compositional mixing via model merging. \citet{chen2025persona} build persona vectors from the model's internal activations for specific traits, enabling deployment-time drift monitoring and post-hoc control via targeted interventions and filtering trait-inducing data. \citet{pai2025billy} further merges multiple persona vectors at inference time for compositional steering in creative generation without additional training. While these methods treat persona vectors as mechanisms for control and monitoring, they do not formulate them as a tool for personality measurement. We take the first step by using persona vectors to quantify LLM personality from internal trait directions, yielding more stable and explainable estimates across prompt variants.

\paragraph{Linear Properties in Hidden-State Space}
Although data live in high-dimensional spaces, their variation often concentrates in a much smaller set of degrees of freedom \citep{Gong:Su:Wang:Wang:You:TKDE23, Ma:Li:Weng:arXiv2510.07703}. This motivates modeling representations with low-dimensional subspaces that are globally nonlinear but locally near linear. Prior work exploits such structure in practice, including rates governed by an effective dimension in deep nets \citep{Suzuki:Nitanda:NeurIPS21}, lightweight adaptation via a few activation-scaling vectors \citep{Liu:Tam:NeurIPS22}, and behavior composition through task vectors in weight space \citep{Ilharco:Ribeiro:Wortsman:Gururangan:2023}. Recent work suggests global behaviors in LLMs vary systematically along low-dimensional directions in hidden-state space \citep{Ilharco:Ribeiro:Wortsman:Gururangan:2023, chen2025persona}. This implies that abstract behavioral traits can be represented as approximately linear directions in the model’s representations \citep{li-zhang-zhang-chen-liu-wang-2025-task-vector}.  We provide the theoretical justification that persona vectors admit an approximately linear structure, and we turn this property into a measurement tool to compute Big Five scores.

\section{Persona-Vector Neutral Interpolation}
\label{sec:big5-geometry}

\begin{figure*}[t]
  \centering
  \includegraphics[width=1.0\textwidth]{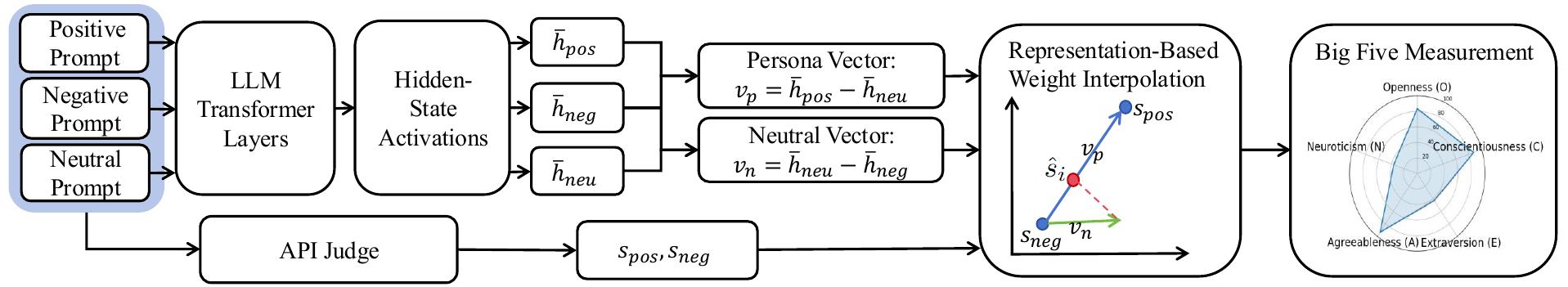}
  \caption{PVNI Pipeline for Prompt-Robust Big Five Trait Measurement. The method extracts a persona direction from positive/negative/neutral prompts, anchors scores on neutral behavior via interpolation and projection, and maps the resulting trait estimates into a stable Big Five subspace.}
  \label{fig:flowchart}
  \vspace{-2mm}
\end{figure*}

As shown in Figure~\ref{fig:flowchart}, we define a representation-space coordinate system for Big Five traits using persona-vector directions, yielding a projection-based estimate. The theoretical justification is developed in Section~\ref{sec:linear-persona-theory}. 

\subsection{Preliminaries and Problem Setup}
\vspace{-1mm}

Let \(M:\mathcal{X}\times\Theta\to\mathcal{Y}\) be a language model with parameters \(\Theta\).
We denote the index layers by \(l\in\{1,\dots,L\}\).
Given an input \(x\) and a prompt \(p\), let $h_l(x,p)\in\mathbb{R}^d$
be the hidden representation at layer $l$ at a fixed probe position.

\paragraph{Persona prompts and representation directions.}
For each trait $i\in\{\mathrm{O},\mathrm{C},\mathrm{E},\mathrm{A},\mathrm{N}\}$, we construct prompt templates $\{p_k^{\,i}\}_{k\in\{\mathrm{pos},\mathrm{neg}, \mathrm{neu}\}}$, define the mean hidden states:
\begin{equation}
\bar h_k^{\,i}
=
\mathbb{E}_{x\in\mathcal{D}}\!\left[h_l\big(x, p_k^{\,i}\big)\right].
\label{eq:mean-hidden}
\end{equation}
We then define the persona vector as
\begin{equation}
v_p^{\,i} = \bar h_{\mathrm{pos}}^{\,i} - \bar h_{\mathrm{neg}}^{\,i},
\label{eq:persona-direction}
\end{equation}
and its unit-normalized version
\begin{equation}
\mu_i \triangleq \frac{v_p^{\,i}}{\|v_p^{\,i}\|_2},\qquad \|\mu_i\|_2=1.
\label{eq:persona-direction-norm}
\end{equation}
Intuitively, $\mu_i$ captures the characteristic direction of trait $i$ in representation space.

\paragraph{Personality trait evaluation.}
For each persona $i$, we assume a collection of inputs $\mathcal{D}=\{x\}$.
Let $h$ denote the layer-$l$ hidden state extracted during generation on input $x$. An API-based judge assigns a score
$r_i(x)\in[0,100]$ to the output on $x$, where higher values indicate stronger expression of persona $i$.
We define the average persona score:
\begin{equation}
\bar s_i(M)=\mathbb{E}_{x\sim\mathcal{D}}\!\left[r_i(x)\right].
\label{eq:avg-score}
\end{equation}

We further define $\hat s_i(M)\in[0,100]$ as the Persona-Vector Neutral Interpolated (PVNI) score estimated by Algorithm~\ref{alg:pv_interpolate_neutral}. We then define the trait axis vector as
\begin{equation}
b_i(M)= \hat s_i(M)\, \mu_i.
\end{equation}


\subsection{Weight Interpolation with Persona Vector}

We estimate a model’s computed neutral score for trait \(i\) via weight interpolation. Specifically, we first obtain two judged score anchors by eliciting trait-promoting and trait-avoiding behaviors with prompts \(p_{\text{pos}}^i\) and \(p_{\text{neg}}^i\). For each \(x\in\mathcal D\), we sample responses \(y_{\text{pos}}^i\) and \(y_{\text{neg}}^i\) and judge them with \(\mathrm{Judge}_i(\cdot)\in[0,100]\), yielding per-example scores \(r_{\text{pos}}^i(x)\) and \(r_{\text{neg}}^i(x)\). We then infer an interpolation weight from hidden-space persona geometry and use it to interpolate between these anchors to obtain the prompt-neutral estimate.

We prompt the judge to output a numeric score. To improve scoring stability, we compute the final trait score as a logit-weighted average over the candidate integer tokens (0--100). For each configuration, we run multiple rollouts and average the resulting scores. We then aggregate across the dataset to obtain two dataset-level anchors:
\begin{equation}
\footnotesize
s_{\text{pos}}^i=\mathbb{E}_{x\sim\mathcal{D}}\!\left[r_{\text{pos}}^i(x)\right], \quad
s_{\text{neg}}^i=\mathbb{E}_{x\sim\mathcal{D}}\!\left[r_{\text{neg}}^i(x)\right].
\end{equation}

Next, we construct persona directions in the model’s hidden space.
Using a hidden-state extractor $\phi(\cdot)$, we obtain representations for the prompted generations and compute mean hidden states $\bar h_{\text{pos}}^i,\bar h_{\text{neg}}^i$.
We additionally collect an explicit neutral prompted condition $p_{\text{neu}}^i$ (used only to locate neutral behavior in representation space) and compute $\bar h_{\text{neu}}^i$.
Consistent with the paper’s direction-extraction practice, we form mean-difference vectors and use response-averaged activations, which empirically yield stronger persona signals than prompt-token alternatives. This gives persona vector and neutral vector:
\begin{equation}
\begin{aligned}
v_p^i &= \bar h_{\text{pos}}^i-\bar h_{\text{neg}}^i,\quad
v_n^i = \bar h_{\text{neu}}^i-\bar h_{\text{neg}}^i .
\end{aligned}
\end{equation}

Finally, we project \(v_n^i\) onto \(v_p^i\) to obtain the interpolation weight:
\begin{equation}
\mathrm{coef}^i
=
\frac{\langle v_{n}^i,\, v_{p}^i \rangle}
     {\langle v_{p}^i,\, v_{p}^i \rangle}.
\end{equation}

We then linearly interpolate between the two score anchors to obtain the estimated prompt-neutral trait score:
\begin{equation}
\hat s_i(M)
=
s_{\text{neg}}^i + \mathrm{coef}^i\cdot\bigl(s_{\text{pos}}^i - s_{\text{neg}}^i \bigr).
\end{equation}

\begin{algorithm}[t]
\caption{Persona-Vector Neutral Interpolation (PVNI) for Original Model Trait Score}
\label{alg:pv_interpolate_neutral}
\begin{algorithmic}
\Require Model $M$, trait $i$, dataset $\mathcal{D}$
\Require Prompts $p_k^i , \forall k\in\{\text{pos},\text{neg},\text{neu}\}$
\Require Judge $\mathrm{Judge}_i(\cdot)\!\in\![0,100]$, extractor $\phi(\cdot)$
\Ensure Prompt-neutral trait score $\hat s_i(M)$

\vspace{0.2em}
\Statex\textbf{Step 1: Responses, scores, hidden states}
\For{$x\in\mathcal{D}$}
  \For{$k\in\{\text{pos},\text{neg},\text{neu}\}$}
    \State $y_k^i \gets M(p_k^i(x))$
    \State $h_k^i(x) \gets \phi(M; p_k^i(x), y_k^i)$
    \If{$k\neq \text{neu}$}
      \State $r_k^i(x)\gets \mathrm{Judge}_i(p_k^i(x),y_k^i)$
    \EndIf
  \EndFor
\EndFor
\State $s_k^i \gets \mathbb{E}_{x\in\mathcal{D}}[r_k^i(x)] \quad \forall k\in\{\text{pos},\text{neg}\}$
\State $\bar h_k^i \gets \mathbb{E}_{x\in\mathcal{D}}[h_k^i(x)] \quad \forall k\in\{\text{pos},\text{neg},\text{neu}\}$

\vspace{0.2em}
\Statex \textbf{Step 2: Projection coefficient in hidden space}
\State $v_p^i \gets \bar h_{\text{pos}}^i-\bar h_{\text{neg}}^i$, \quad
       $v_n^i \gets \bar h_{\text{neu}}^i-\bar h_{\text{neg}}^i$
\State $\mathrm{coef^i}\gets \dfrac{\langle v_n^i, v_p^i\rangle}{\langle v_p^i, v_p^i\rangle}$;\;
       $\mathrm{coef^i}\gets \mathrm{Clip}(\mathrm{coef^i},0,1)$

\vspace{0.2em}
\Statex \textbf{Step 3: Neutral score by interpolation}
\State \Return $\hat s_i(M)\gets s_{\text{neg}}^i +\mathrm{coef^i}\,(s_{\text{pos}}^i-s_{\text{neg}}^i)$
\vspace{-1mm}
\end{algorithmic}
\end{algorithm}

Algorithm~\ref{alg:pv_interpolate_neutral} provides the pseudocode for PVNI.
We apply Algorithm~\ref{alg:pv_interpolate_neutral} independently to each of the Big Five traits
$i \in \{\textsc{O},\textsc{C},\textsc{E},\textsc{A},\textsc{N}\}$,
using trait-specific prompt pairs $(p_{\text{pos}}^{i},p_{\text{neg}}^{i})$ and an auxiliary neutral prompt
$p_{\text{neu}}^{i}$ only for locating neutral behavior in representation space.
For each trait $i$, the procedure returns an estimated computed neutral score
$\hat s_i(M)\in[0,100]$.

Collecting the five neutral scores yields a stable, relatively prompt-neutral Big Five coordinate vector for model $M$: 

\begin{equation}
\small
\hat s(M) \;=\;
\bigl[\hat s_O(M),\ldots,\hat s_N(M)\bigr]^\top
\;\in\; \mathbb{R}^{5},
\label{eq:big5_coord_vector}
\end{equation}

\vspace{-2mm}

\begin{equation}
\small
B(M)=\bigl[\hat s_O(M)\mu_O,\ldots,\hat s_N(M)\mu_N\bigr]
\;\in\;\mathbb{R}^{d\times 5},
\label{eq:B_matrix}
\end{equation}
\vspace{-2mm}

We interpret $B(M)$ as the model's Big Five (OCEAN) personality embedding---Openness, Conscientiousness, Extraversion, Agreeableness, and Neuroticism. 
Each axis is anchored by a trait-promoting direction and mapped to a neutral score via weight interpolation.

\section{Linear Theory of Persona Vector}
\label{sec:linear-persona-theory}

PVNI builds on the observed linearity of persona directions. This section proves that persona-induced representation changes are approximately linear, providing the theoretical basis for representation-based weight interpolation as a vector operation for personality trait evaluation in Section~\ref{sec:big5-geometry}.

\paragraph{Persona loss.} We previously defined the average persona score as
\begin{equation}
\bar{s}_i(M)= \bar{s}_i(h)
= \mathbb{E}_{x\sim D_i}\!\left[\, r_i(x; h) \,\right],
\end{equation}
where $D_i$ denotes the input prompt set associated with persona $i$. After normalizing scores to $[0,1]$, we assume that applying the persona update yields a near-maximal score:
\begin{equation}
\bar{s}_i\!\big(h+v_p^i\big)\ge 1-\varepsilon,
\qquad \text{for a small }\varepsilon>0.
\label{eq:persona-evaluated}
\end{equation}

We define the normalized persona loss as the complement of the normalized score:
\begin{equation}
\mathcal{L}_i(h)\triangleq 1-\bar{s}_i(h)\in[0,1].
\label{eq:persona-loss}
\end{equation}
Under this definition, the near-maximal-score condition in~\eqref{eq:persona-evaluated}
is equivalent to the following small-loss condition:
\begin{equation}
\mathcal{L}_i\!\big(h+v_p^i\big)\le \varepsilon.
\label{eq:persona-loss-eps}
\end{equation}

\paragraph{Persona correlation.}
For two personas $i_1,i_2$, we define their representation-space correlation as
\begin{equation}
\alpha(i_1,i_2)\triangleq \mu_{i_1}^{\top}\mu_{i_2}\in[-1,1].
\label{eq:persona-corr}
\end{equation}
We say that $i_1$ and $i_2$ are \emph{aligned} if $\alpha(i_1,i_2)>0$,
\emph{contradictory} if $\alpha(i_1,i_2)<0$,
and \emph{orthogonal} if $\alpha(i_1,i_2)=0$.

\subsection{Representation and Local Linearity}
\label{subsec:rep-model}

We work with a stylized representation model capturing the empirical observation that persona manipulations act primarily along a low-dimensional trait subspace.

\begin{assumption}[Local Linearity of Persona Scores]
\label{ass:local-linearity}
For each persona $i$, there exist a score function $g_i:\mathbb{R}^d\to\mathbb{R}$,
constants $a_i>0$, $L_i>0$, and $r_i>0$ such that for any typical hidden state $h$ at layer $\ell$
and any perturbation $\delta\in\mathcal{U}$ with $\|\delta\|_2\le r_i$,
\begin{equation}
\label{eq:local-linear-gp}
g_i(h+\delta)
=
g_i(h) + a_i\langle \delta,\mu_i\rangle + \varepsilon_i(h,\delta).
\end{equation}

\begin{equation}
\label{eq:local-linear-err}
|\varepsilon_i(h,\delta)|
\le
L_i\|\delta\|_2^2.
\end{equation}
Moreover, the expected persona loss $\mathcal{L}_i(h)$ depends on $h$ only through the
distribution of $g_i\!\big(h_\ell(x,p_k^{\,i})\big)$ over $(x,k)\in\mathcal{D}_i\times\{\mathrm{pos},\mathrm{neg},\mathrm{neu}\}$,
and is non-increasing in the mean margin
$\mathbb{E}_{(x,k)}\!\left[g_i\!\big(h_\ell(x,p_k^{\,i})\big)\right]$.
\end{assumption}

\begin{assumption}[Well-Trained Persona Adaptation]
\label{ass:well-trained}
For each persona $i$, the persona-adapted parameters
$\Theta_i^* \triangleq \Theta^{(0)}+\Delta\Theta_i$
induce at layer $l$ the activation shift
\begin{equation}\label{eq:delta-h-def}
\footnotesize
\Delta h_l^{\,i}(x,p)
=
h_l(x,p;\Theta^{(0)}+\Delta\Theta_i)-h_l(x,p;\Theta^{(0)}).
\end{equation}
On typical $(x,p)$, there exists $c_i>0$ such that
\begin{equation}
\footnotesize
\Delta h_\ell^{\,i}(x,p)
\;=\;
c_i\,\big\langle h_\ell(x,p;\Theta^{(0)}),\mu_i\big\rangle\,\mu_i
\;+\; r_i(x,p),
\label{eq:align-structure}
\end{equation}
with $\|r_i(x,p)\|_2 \le \beta\,\|h_\ell(x,p;\Theta^{(0)})\|_2$ for some small $\beta>0$.

Letting $V_\ell\in\mathbb{R}^{m\times d}$ denote the MLP weight matrix at layer $\ell$
and $\Delta V_\ell^{\,i}$ the persona-induced update, $\exists\,\mathcal{S}_i \subseteq [m]\ \text{with}\ |\mathcal{S}_i|\ge \rho m,\ \exists\,0<c<C\ \text{s.t.}\ \forall t\in[m]$
\begin{equation}
\label{eq:mlp-row-structure}
\footnotesize
\begin{aligned}
\begin{cases}
\|\Delta V_{\ell,t}^{\,i}\|_2 \ge c\,m^{-1/2},\ \ \Delta V_{\ell,t}^{\,i}\approx \gamma_{i,t}\mu_i, & t\in\mathcal S_i,\\[2pt]
\|\Delta V_{\ell,t}^{\,i}\|_2 \le C\,\dfrac{\sqrt{\log m}}{m}, & t\notin\mathcal S_i.
\end{cases}
\end{aligned}
\end{equation}
where $\gamma_{i,t}\in\mathbb{R}$ are scalar coefficients capturing the approximate alignment with $\mu_i$.
\end{assumption}

Under the above assumptions, persona vectors admit a simple approximate structure.

\begin{lemma}[Persona vectors as approximate rank-one amplifiers]
\label{lem:persona-structure}
Assume Assumptions~\ref{ass:local-linearity}--\ref{ass:well-trained}.
Then for each persona $i$, there exist a direction $\mu_i$ and a constant $c_i>0$ such that,
for all hidden states $h$ in the typical region at layer $\ell$,
\begin{equation}
\Delta h_\ell^{\,i}(h)
=
c_i\,\langle h,\mu_i\rangle\,\mu_i
\;+\;
e_i(h),
\label{eq:rank-one-form}
\end{equation}
and the residual satisfies
\begin{equation}
\|e_i(h)\|_2 \le \beta \|h\|_2.
\label{eq:rank-one-residual}
\end{equation}
Moreover,~\eqref{eq:rank-one-form} is dominated by the sparse MLP row set $\mathcal{S}_i$:
pruning $\mathcal{S}_i^{c}$ changes $\Delta h_\ell^{\,i}(h)$ by at most $O(\beta\|h\|_2)$.
The induced attention reweighting is aligned with $\mu_i$, amplifying interactions with
$\langle h,\mu_i\rangle>0$ and suppressing the rest up to $O(\beta)$.
\end{lemma}

\noindent
Lemma~\ref{lem:persona-structure} shows that persona vectors act as \emph{directional amplifiers} that selectively boost the \(\mu_i\)-component of the hidden state, yielding near-linear additivity in persona editing.

\subsection{Multi-Persona Composition and Negation}
\label{subsec:composition}

We now study how two persona vectors interact under linear composition.
For two traits \(i,j\in\mathcal{I}\) with \(i\neq j\), consider the one-parameter family of shifted hidden states
\begin{equation}
h(\lambda)
\;\triangleq\;
h + v_p^{\,i} + \lambda\,v_p^{\,j},
\qquad \lambda\in\mathbb{R}.
\label{eq:two-persona-model}
\end{equation}

\begin{theorem}[Multi-Persona Composition]
\label{thm:multi-persona}
Let \(i,j\in\mathcal{I}\) be two traits, and write \(\alpha=\alpha(i,j)=\mu_i^\top\mu_j\).
Suppose Assumptions~\ref{ass:local-linearity}--\ref{ass:well-trained} hold, and
\(\mathcal{L}_{i}(h+v_p^{\,i})\le \varepsilon\) and \(\mathcal{L}_{j}(h+v_p^{\,j})\le \varepsilon\).
Then there exists a constant \(C>0\) such that:
\begin{enumerate}
\item If \(\alpha\ge 0\), then for every
\begin{equation}
\lambda \;\ge\; 1-\alpha + C\beta,
\end{equation}
\vspace{-1mm}
we have
\begin{equation}
\mathcal{L}_{i}\!\big(h(\lambda)\big)
\;\le\; \mathcal{O}(\varepsilon) + \mathcal{O}(|\lambda|\beta),
\end{equation}
and
\begin{equation}
\mathcal{L}_{j}\!\big(h(\lambda)\big)
\;\le\; \mathcal{O}(\varepsilon) + \mathcal{O}(\beta).
\end{equation}

\item If \(\alpha<0\), then for any \(\lambda\), at least one of
\(\mathcal{L}_{i}\!\big(h(\lambda)\big)\) or \(\mathcal{L}_{j}\!\big(h(\lambda)\big)\)
is bounded below by a constant independent of \(\varepsilon\).
\end{enumerate}
\end{theorem}

\begin{remark}
When \(i\), \(j\) are orthogonal (\(\alpha(i,j)=0\)), Theorem~\ref{thm:multi-persona}
implies that a constant-scale coefficient \(\lambda\) suffices to express both traits simultaneously.
Positive correlation (\(\alpha(i,j)>0\)) reduces the required scale, whereas negative correlation
(\(\alpha(i,j)<0\)) induces an inherent trade-off: no single \(\lambda\) can keep both
\(\mathcal{L}_{i}\!\big(h(\lambda)\big)\) and \(\mathcal{L}_{j}\!\big(h(\lambda)\big)\) small.
\end{remark}

\vspace{1mm}

\begin{theorem}[Persona Negation]
\label{thm:persona-unlearning}
Under the assumptions of Theorem~\ref{thm:multi-persona}, there exist universal constants
\(c_0,c_1,c_2>0\) such that:
\begin{enumerate}
\item \textbf{Orthogonal traits (\(\alpha(i,j)=0\)).}
For all \(\lambda \le -c_1\),
\begin{equation}
\mathcal{L}_{i}\!\big(h(\lambda)\big)
\;\le\; \mathcal{O}(\varepsilon) + \mathcal{O}(|\lambda|\beta),
\end{equation}
and
\begin{equation}
\mathcal{L}_{j}\!\big(h(\lambda)\big)
\;\ge\; c_0.
\end{equation}
That is, negatively scaling \(v_p^{\,j}\) suppresses trait \(j\) while preserving trait \(i\).

\item \textbf{Contradictory traits (\(\alpha(i,j)<0\)).}
There exists an interval
\[
I_\alpha
\;=\;
\big[-c_2/\alpha^2,\; c_2/|\alpha|\big]
\]
such that for all \(\lambda \in I_\alpha\),
\begin{equation}
\mathcal{L}_{i}\!\big(h(\lambda)\big)
\;\le\; \mathcal{O}(\varepsilon) + \mathcal{O}(|\lambda|\beta),
\end{equation}
and
\begin{equation}
\mathcal{L}_{j}\!\big(h(\lambda)\big)
\;\ge\; c_0.
\end{equation}
Thus, when \(\alpha(i,j)<0\), there is a non-trivial range of negative coefficients that deletes \(j\) while preserving \(i\).

\item \textbf{Mildly aligned traits (\(0<\alpha(i,j)<1-c_0\)).}
There exists \(c_3>0\) such that for all \(\lambda \in [0,c_3]\),
\begin{equation}
\mathcal{L}_{i}\!\big(h(\lambda)\big)
\;\le\; \mathcal{O}(\varepsilon) + \mathcal{O}(\beta),
\end{equation}
and
\begin{equation}
\mathcal{L}_{j}\!\big(h(\lambda)\big)
\;\ge\; \Omega(\alpha)\;-\;\mathcal{O}(\beta).
\end{equation}
In this regime, small positive coefficients can weaken trait \(j\) without destroying trait \(i\).
\end{enumerate}
\end{theorem}

\begin{remark}
Theorem~\ref{thm:persona-unlearning} shows that negation is easiest for orthogonal or contradictory traits and becomes harder as \(\alpha(i,j)\) increases.
Empirically, subtraction (\(\lambda<0\)) is effective for antagonistic traits but has limited impact when the two traits are highly aligned.
\end{remark}

\subsection{Persona Linear Subspaces Generalization}
\label{subsec:persona-ood}

We now consider synthesizing an out-of-domain persona from a collection of in-domain trait directions.
Let \(\mathcal{I}= \{\mathrm{O},\mathrm{C},\mathrm{E},\mathrm{A},\mathrm{N}\}\) denote the Big Five trait index set.
Assume the corresponding unit directions \(\{\mu_i\}_{i\in\mathcal{I}}\) form an orthonormal basis of a trait subspace \(\mathcal{U}\subset\mathbb{R}^d\).
Let \(i^\star\) denote an out-of-domain persona with unit direction \(\mu_{i^\star}\).

The direction \(\mu_{i^\star}\) is decomposed into its component in \(\mathcal{U}\) and an orthogonal residual:
\begin{equation}
\footnotesize
\mu_{i^\star}
\;=\;
\sum_{i\in\mathcal{I}} \gamma_i \mu_i
\;+\;
\kappa\,\mu_\perp,
\quad
\mu_\perp \perp \mu_i,\ \ \forall i\in\mathcal{I},
\label{eq:persona-decomp}
\end{equation}
where \(\gamma_i\in\mathbb{R}\) are coefficients and \(\mu_\perp\) is a residual direction.
We assume \(|\kappa|\le \kappa_0\) for some small constant \(\kappa_0\).
Persona arithmetic is then performed by a linear combination of the in-domain persona updates \(\{v_p^{\,i}\}_{i\in\mathcal{I}}\).
For \(\lambda=(\lambda_i)_{i\in\mathcal{I}}\in\mathbb{R}^{|\mathcal{I}|}\), define
\begin{equation}
h(\lambda)
\;\triangleq\;
h + \sum_{i\in\mathcal{I}} \lambda_i\, v_p^{\,i}.
\label{eq:persona-ood-arith}
\end{equation}

\begin{theorem}[Out-of-Domain Persona Synthesis]
\label{thm:persona-ood}
Suppose Assumptions~\ref{ass:local-linearity}--\ref{ass:well-trained} hold for all \(i\in\mathcal{I}\) and for \(i^\star\),
and that \(|\kappa|\le \kappa_0\) in~\eqref{eq:persona-decomp}.
Then there exists \(c\in(0,1)\) and constants \(C_1,C_2>0\) such that if the coefficients \(\lambda\) satisfy
\begin{equation}
\small
\sum_{i\in\mathcal{I}} \lambda_i \gamma_i \,\ge\, 1 + c,
\quad
\sum_{i\in\mathcal{I}} \lambda_i \gamma_i^2 \,\ge\, 1 + c,
\quad
|\lambda_i|\,\beta \,\le\, C_1 c\ ,
\label{eq:lambda-gamma-cond}
\end{equation}
then
\begin{equation}
\mathcal{L}_{i^\star}\!\big(h(\lambda)\big)
\;\le\;
\mathcal{O}(\varepsilon)
\;+\;
\mathcal{O}\!\big(\beta + \kappa_0^2\big).
\label{eq:persona-ood-loss}
\end{equation}
\end{theorem}
\vspace{-1mm}

\begin{remark}
Conditions~\eqref{eq:lambda-gamma-cond} ensure sufficient combined margin along the in-subspace component of \(\mu_{i^\star}\),
while keeping off-direction residuals controlled.
Thus, when an out-of-domain persona direction lies mostly in \(\mathrm{span}\{\mu_i\}_{i\in\mathcal{I}}\),
it can be synthesized by a suitable linear combination of \(\{v_p^{\,i}\}_{i\in\mathcal{I}}\).
\end{remark}

\section{Experiments}
For each trait $i\in\{\textsc{O},\textsc{C},\textsc{E},\textsc{A},\textsc{N}\}$, we run Algorithm~\ref{alg:pv_interpolate_neutral} once to obtain a neutral score $\hat s_i(M)\in[0,100]$. The resulting Big Five coordinate vector is $\small B(M)\triangleq\bigl[\hat s_O(M)v_O,\ldots,\hat s_N(M)v_N\bigr]$.

\begin{table*}[t]
\centering
\setlength{\tabcolsep}{5pt}
\renewcommand{\arraystretch}{1.0}

\begin{threeparttable}
\begin{adjustbox}{max width=\textwidth}
\begin{tabular}{lllcccc}
\toprule
\textbf{What Varies} & \textbf{Model} & \textbf{Trait} &
\multicolumn{2}{c}{\textbf{Self Report Assessment}} &
\textbf{Open-ended Elicitation} &
\textbf{PVNI (Ours)} \\
\cmidrule(lr){4-5}
& & & \textbf{IPIP-BFFM-50} & \textbf{IPIP-NEO-120} & & \\
\midrule

\multirow{10}{*}{\shortstack{Questionnaire\\variants}}
& \multirow{5}{*}{Qwen-2.5-7B}
 & Openness (O)          & 82.5 $\pm$ 6.40  & 87.5 $\pm$ 3.74  & 82.49 $\pm$ 1.09  & 83.55 $\pm$ \textbf{0.82} \\
&  & Conscientiousness (C) & 60.0 $\pm$ 7.20  & 72.9 $\pm$ 5.31  & 90.52 $\pm$ 2.53  & 87.63 $\pm$ \textbf{0.73} \\
&  & Extraversion (E)      & 47.5 $\pm$ 10.50 & 66.7 $\pm$ 12.76 & 58.82 $\pm$ 15.31 & 42.89 $\pm$ \textbf{2.49} \\
&  & Agreeableness (A)     & 75.0 $\pm$ 5.90  & 88.5 $\pm$ 4.21  & 93.31 $\pm$ 0.69  & 93.39 $\pm$ \textbf{0.68} \\
&  & Neuroticism (N)       & 45.0 $\pm$ 18.10 & 38.5 $\pm$ 8.98  & 28.79 $\pm$ 2.88  & 36.45 $\pm$ \textbf{0.83} \\
\cmidrule(lr){2-7}
& \multirow{5}{*}{Llama-3-8B}
 & Openness (O)          & 72.5 $\pm$ 5.80  & 86.0 $\pm$ 3.43  & 96.23 $\pm$ 1.93  & 94.12 $\pm$ \textbf{0.74} \\
&  & Conscientiousness (C) & 62.5 $\pm$ 6.90  & 77.7 $\pm$ 5.21  & 97.68 $\pm$ 1.88  & 86.58 $\pm$ \textbf{0.51} \\
&  & Extraversion (E)      & 57.5 $\pm$ 9.80  & 74.0 $\pm$ 7.05  & 39.28 $\pm$ 4.43  & 51.02 $\pm$ \textbf{1.27} \\
&  & Agreeableness (A)     & 77.5 $\pm$ 5.10  & 85.4 $\pm$ 2.17  & 95.32 $\pm$ 0.54  & 95.84 $\pm$ \textbf{0.38} \\
&  & Neuroticism (N)       & 42.5 $\pm$ 10.40 & 30.6 $\pm$ 8.92  & 12.06 $\pm$ 3.10  & 32.07 $\pm$ \textbf{1.13} \\

\midrule

\multirow{10}{*}{\shortstack{Role-play\\variants}}
& \multirow{5}{*}{Qwen-2.5-7B}
 & Openness (O)          & 82.3 $\pm$ 4.10  & 87.8 $\pm$ 2.63  & 82.61 $\pm$ 0.72  & 83.42 $\pm$ \textbf{0.55} \\
&  & Conscientiousness (C) & 60.4 $\pm$ 4.80  & 72.5 $\pm$ 3.68  & 90.14 $\pm$ 1.63  & 87.92 $\pm$ \textbf{0.50} \\
&  & Extraversion (E)      & 47.1 $\pm$ 7.20  & 66.9 $\pm$ 8.91  & 59.18 $\pm$ 9.83  & 43.10 $\pm$ \textbf{1.60} \\
&  & Agreeableness (A)     & 75.4 $\pm$ 3.90  & 88.1 $\pm$ 2.97  & 93.02 $\pm$ 0.45  & 93.58 $\pm$ \textbf{0.42} \\
&  & Neuroticism (N)       & 44.6 $\pm$ 11.20 & 38.9 $\pm$ 6.24  & 28.52 $\pm$ 2.17  & 36.18 $\pm$ \textbf{0.60} \\
\cmidrule(lr){2-7}
& \multirow{5}{*}{Llama-3-8B}
 & Openness (O)          & 72.6 $\pm$ 3.90  & 86.1 $\pm$ 2.65  & 96.34 $\pm$ 1.35  & 94.05 $\pm$ \textbf{0.58} \\
&  & Conscientiousness (C) & 62.4 $\pm$ 4.60  & 77.8 $\pm$ 3.83  & 97.59 $\pm$ 1.12  & 86.63 $\pm$ \textbf{0.44} \\
&  & Extraversion (E)      & 57.6 $\pm$ 6.10  & 74.1 $\pm$ 5.20  & 39.41 $\pm$ 3.16  & 51.10 $\pm$ \textbf{0.98} \\
&  & Agreeableness (A)     & 77.4 $\pm$ 3.50  & 85.3 $\pm$ 1.76  & 95.18 $\pm$ 0.41  & 95.77 $\pm$ \textbf{0.30} \\
&  & Neuroticism (N)       & 42.6 $\pm$ 7.40  & 30.7 $\pm$ 6.11  & 12.18 $\pm$ 2.24  & 32.02 $\pm$ \textbf{0.86} \\

\bottomrule
\end{tabular}
\end{adjustbox}
\end{threeparttable}

\caption{Big Five (OCEAN) personality ratings across similarly-sized LLMs under different evaluation protocols.
Results are shown as mean $\pm$ std across questionnaire/role-play variants.
In the PVNI (Ours) column, the standard deviation term is \textbf{boldfaced} since PVNI consistently achieves the lowest variability among all methods.}
\label{tab:big5_rating_comparison}
\vspace{-2mm}
\end{table*}

\begin{figure}[t]
  \centering
  \includegraphics[width=\linewidth]{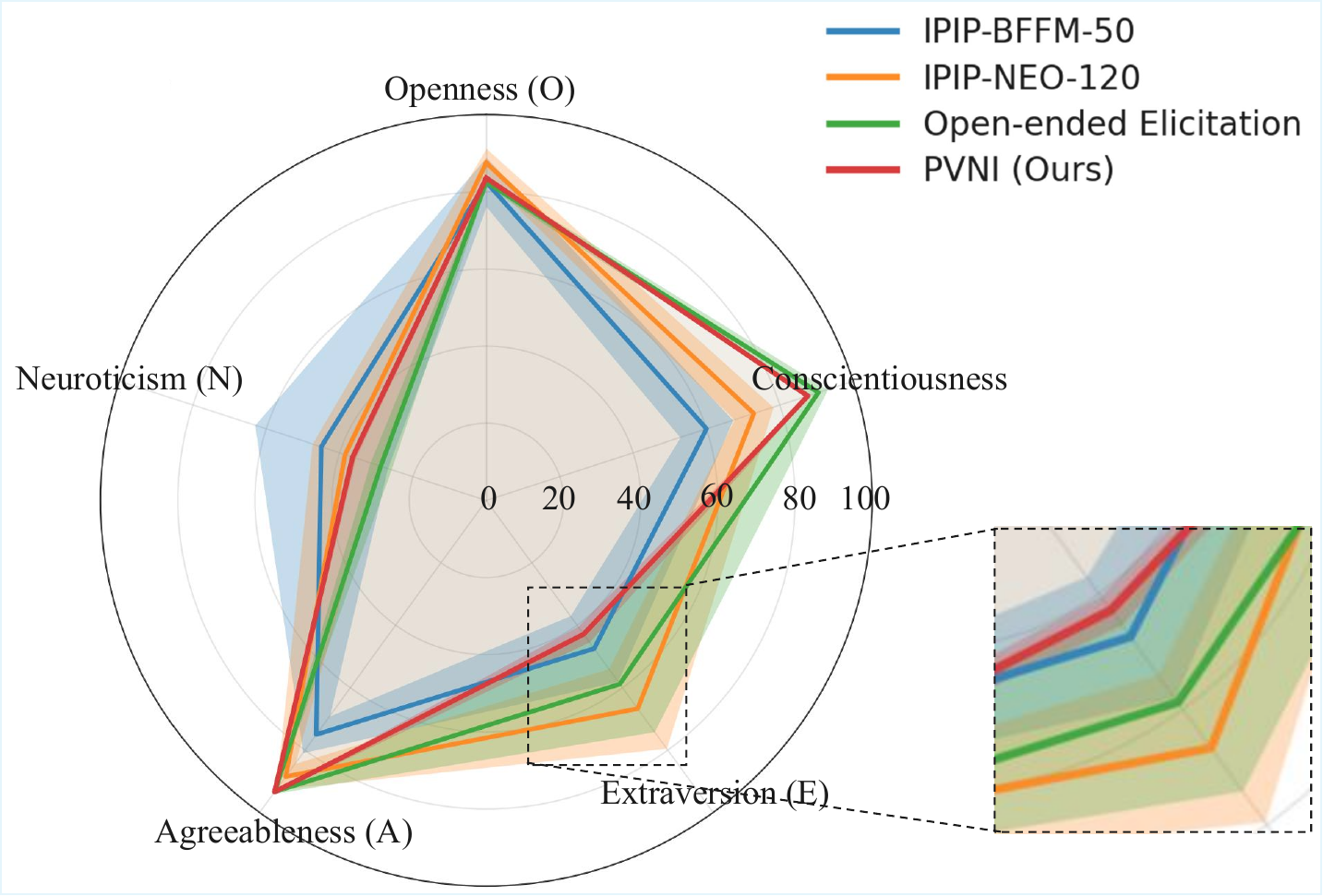}
  \caption{Big Five trait radar plots under four evaluation protocols across Qwen-2.5-7B. Shaded bands indicate standard deviation over questionnaire variants.}
  \label{fig:radar-one}
  \vspace{-2mm}
\end{figure}

\begin{figure*}[t]
  \centering
  \captionsetup{skip=3pt}
  \captionsetup[subfigure]{justification=centering,singlelinecheck=false}

  \begin{subfigure}[t]{0.49\textwidth}
    \centering
    \includegraphics[width=\linewidth]{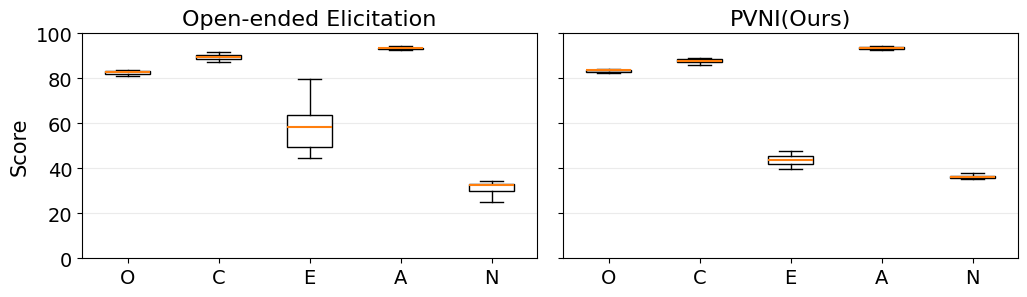}
    \caption{Questionnaire variants for Qwen-2.5-7B}
    \label{fig:box_qnn_qwen}
  \end{subfigure}\hfill
  \begin{subfigure}[t]{0.49\textwidth}
    \centering
    \includegraphics[width=\linewidth]{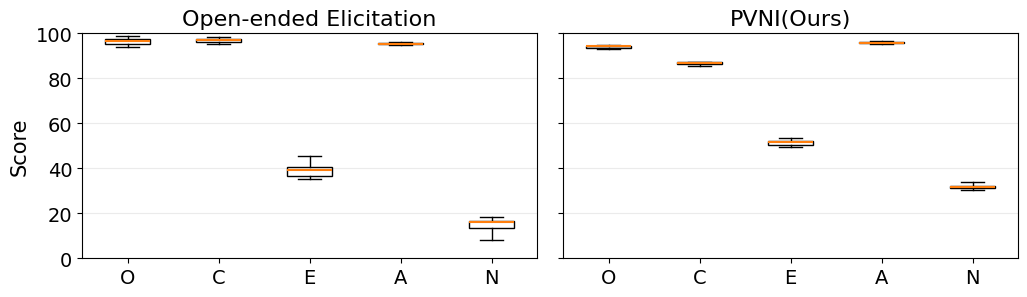}
    \caption{Questionnaire variants for Llama-3-8B}
    \label{fig:box_qnn_llama}
  \end{subfigure}

  \par\medskip 

  \begin{subfigure}[t]{0.49\textwidth}
    \centering
    \includegraphics[width=\linewidth]{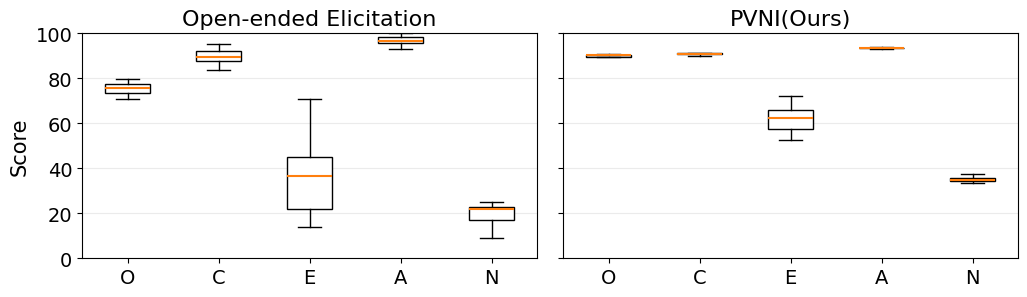}
    \caption{Questionnaire variants for Mistral-7B-v0.1}
    \label{fig:box_qnn_mistral}
  \end{subfigure}\hfill
  \begin{subfigure}[t]{0.49\textwidth}
    \centering
    \includegraphics[width=\linewidth]{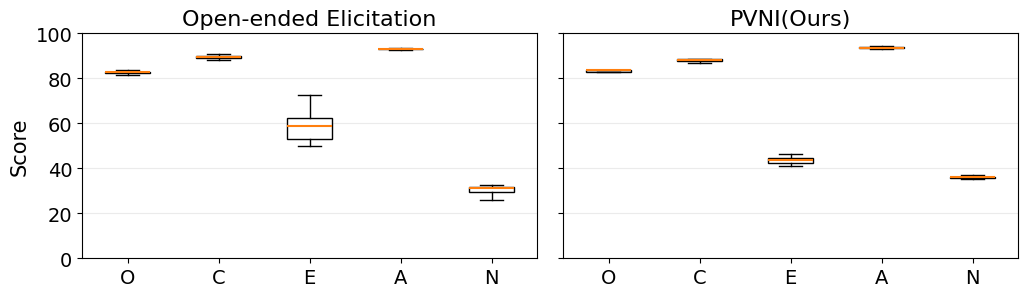}
    \caption{Role-play variants for Qwen-2.5-7B}
    \label{fig:box_rp_qwen}
  \end{subfigure}

  \par\medskip

  \begin{subfigure}[t]{0.49\textwidth}
    \centering
    \includegraphics[width=\linewidth]{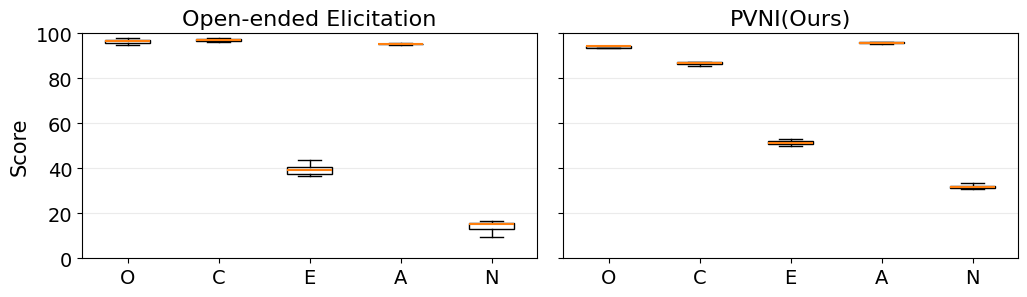}
    \caption{Role-play variants for Llama-3-8B}
    \label{fig:box_rp_llama}
  \end{subfigure}\hfill
  \begin{subfigure}[t]{0.49\textwidth}
    \centering
    \includegraphics[width=\linewidth]{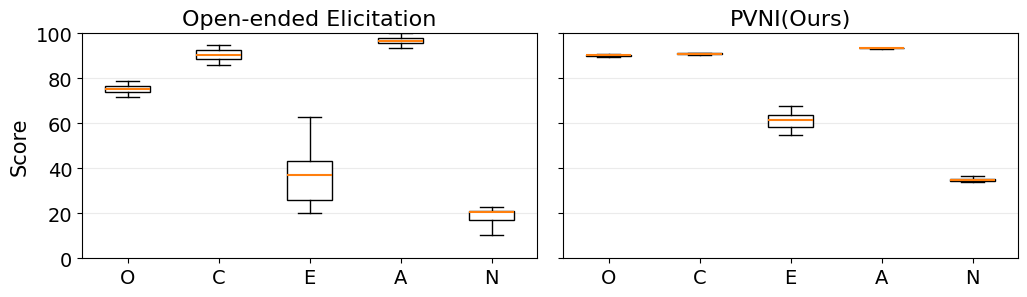}
    \caption{Role-play variants for Mistral-7B-v0.1}
    \label{fig:box_rp_mistral}
  \end{subfigure}

  \caption{Boxplots under questionnaire and role-play variants for Qwen-2.5-7B, Llama-3-8B, and Mistral-7B-v0.1.}
  \label{fig:box_3x2_variants}
  \vspace{-2mm}
\end{figure*}

\subsection{Experimental Settings}

\paragraph{Test Models.}
We conduct Big Five personality trait evaluation on three open-source LLMs: Qwen-2.5-7B-Instruct \citep{QwenTeam:Qwen25:2024}, Llama-3-8B-Instruct \citep{MetaAI:Llama3ModelCard:2024}, and Mistral-7B-v0.1-Instruct \citep{Jiang:Sablayrolles:Mensch:etAl:arXiv2310.06825}. For each model $M$, we estimate trait coordinates for $i\!\in\!\{\textsc{O},\textsc{C},\textsc{E},\textsc{A},\textsc{N}\}$ using Algorithm~\ref{alg:pv_interpolate_neutral}, and report the resulting neutral prompt scores $\hat s_i(M)$ as the model’s Big Five personality level.

\paragraph{Judge API and Artifact Generation.}
Trait evaluation uses an external judge API (GPT-4.1-mini) that outputs an open-ended elicitation score in $[0,100]$ per response. We use GPT-5.2 to generate contrastive system prompts and trait-eliciting questions, split into extraction and evaluation sets.

\paragraph{Variants Construction.} We use two controlled prompt variants for each protocol as shown in Appendix~\ref{sub:variant}. For Self Report Assessment, the questionnaire variant rewrites same questions with equivalent wording, while role-play variant adds one brief role line and keeps the questions unchanged. For Open-ended Elicitation/PVNI, questionnaire variant uses different questions with same trait direction, while the role-play variant rewrites the pos/neg instruction. All other settings are fixed, so variation reflects prompting not semantics.

\paragraph{Hardware and Runtime.}
Experiments were run on NVIDIA RTX 4090 GPUs. Personality measurement pipeline covers all five traits and includes generation, judging, and representation extraction, taking about 2 GPU-hours per model. 

\subsection{Big Five Personality Trait Evaluation}

We evaluate Qwen-2.5-7B, Llama-3-8B, and Mistral-7B-v0.1 on Big Five using IPIP-BFFM-50, IPIP-NEO-120, Open-ended Elicitation, and PVNI. Each protocol uses 10 prompt sets for questionnaire and role-play variants. We report mean ± std across sets to quantify prompt robustness.

As shown in Table~\ref{tab:big5_rating_comparison} and Table~\ref{tab:big5_rating_comparison_combined}, PVNI is the most stable method. It consistently achieves the smallest standard deviation across all models and traits, showing minimal prompt dependence and the strongest robustness to prompt rewrites.

Across protocols, questionnaire and role-play variants yield similar means, implying the estimated trait levels are largely unchanged by the variant type. However, role-play variants typically have slightly lower variance, likely because questionnaire variants modify the question content more substantially, introducing larger perturbations.

We also observe a clear mean–variance coupling across traits: traits with lower mean scores (e.g. Neuroticism and Extraversion) tend to exhibit larger standard deviations, indicating higher uncertainty when the trait strength is weak.

In Figure~\ref{fig:radar-one}, PVNI (Ours) shows the narrowest shaded uncertainty bands across all five traits. Since the shading denotes $\pm$ one standard deviation over questionnaire variants, the smaller band area indicates lower variance. This trend is consistent across all three models (Qwen, Llama, and Mistral) in Appendix Figure~\ref{fig:radar} under both questionnaire and role-play variants. Thus, PVNI remains robust to rephrasing, while other protocols exhibit wider bands that reflect stronger prompt-induced fluctuations.

\subsection{Protocol-wise Variability}
\label{subsec:protocol_variability}

Figure~\ref{fig:box_3x2_variants} shows that Open-ended Elicitation produces wider boxes and longer whiskers across three LLMs, indicating strong sensitivity to prompt variants and unstable Big Five estimates, whereas PVNI yields consistently tighter distributions with small IQRs and short whiskers. Appendix Figure~\ref{fig:boxplots_4methods} further confirms this trend in a full comparison: both self-report questionnaire protocols, IPIP-BFFM-50 and IPIP-NEO-120, exhibit larger spread and higher variability than PVNI across traits and models, making PVNI the most robust and stable measurement under rephrasing.

\section{Conclusion}
\vspace{-1mm}

We proposed Persona-Vector Neutrality Interpolation (PVNI), an internal-activation-based weight interpolation method for stable and explainable personality trait evaluation in LLMs. PVNI extracts trait directions as persona vectors from contrastive prompts, then estimates prompt-neutral trait strength via projection onto the axis and interpolation between judged anchors. We also develop a linear theory that justifies these operations. Across LLMs and both questionnaire and role-play variants, PVNI consistently reduces variance over prior protocols. Future work will extend beyond Big Five traits, reduce reliance on external judges, and test robustness across broader domains and languages.

\clearpage

\section{Limitations}
We structure our limitations section as arguments and counterarguments, inspired by \citet{balepur-etal-2025-best}:

\paragraph{Judging can introduce bias:}
PVNI uses a judge to obtain pos/neg anchor scores, so absolute values can reflect judge preferences. However, PVNI uses the judge only for anchoring, while the interpolation weight is computed from the model's internal representation geometry. In practice, PVNI’s main claim is variance reduction across prompt variants, not judge-invariant absolute calibration. Future work includes multi-judge ensembles, human calibration, and distilling a stable judge.

\paragraph{Neutrality may not lie on the pos--neg axis:}
Projection-based interpolation assumes the neutral condition lies approximately along the pos--neg axis. This can be violated when trait effects are curved, multi-modal, or strongly context-dependent. We reduce pathological behavior by clipping the interpolation weight and averaging across prompt sets, but this does not guarantee perfect calibration in highly nonlinear regions. Extending PVNI to multi-anchor or nonlinear calibration is an important direction.

\paragraph{PVNI requires white-box access:}
PVNI requires hidden-state extraction to compute persona directions and interpolation weights. This limits direct applicability to closed, API-only models. Our focus is research-grade, representation-based evaluation with transparent geometric operations. A practical next step is to study accessible surrogates, such as logit-space proxies or distillation to a model with representation access.

\paragraph{Trait directions are not independent:}
Persona directions across traits can be correlated, so shifting one trait may partially move others. This complicates axis-specific interpretation and can blur cross-trait comparisons. PVNI can measure these correlations and treats traits as a shared low-dimensional structure rather than perfectly independent axes. Future work can incorporate constrained subspace learning or joint multi-trait estimation to better disentangle directions.

\paragraph{Our evaluation does not cover all settings:}
Experiments focus on Big Five traits, a small set of open-source instruction-tuned models, and controlled prompt variants, largely in a single-turn format. The results may not fully transfer to multilingual settings, long conversations, or tool-augmented agents. The controlled design is intentional for isolating prompt robustness effects, but broader coverage remains needed. Future work should test more languages, domains, model families, and multi-turn consistency.

\paragraph{Prompt ensembles may be incomplete:}
Persona directions are extracted from designed contrastive prompts and elicitation contexts. Even if estimates are stable under controlled variants, they can still be incomplete proxies for the underlying trait if prompt space coverage is narrow. We partially mitigate this by using prompt sets and two variant types while holding all other settings fixed. Expanding prompt ensembles and using adversarial or active prompt selection would strengthen coverage.

\paragraph{Stability comes with overhead:}
PVNI involves generation, judging, and activation extraction for multiple traits and prompt sets. This can be non-trivial at scale, especially for larger models. The pipeline is parallelizable and can benefit from caching and reuse, but cost remains a practical constraint. Future work should report cost--stability trade-offs and explore sample-efficient designs.

\bibliography{references}

\appendix

\clearpage
\onecolumn
\appendix

\section{Appendix}
\label{sec:appendix}

\subsection{Questionnaire and Role-Play Variants}
\label{sub:variant}

\begin{promptbox}{Self Report with Role-Play Variants}
\label{box:ipiproleplay}
\begin{enumerate}
  \item Respond as an LLM agent whose answers reflect a coherent personality.
  \item Adopt a consistent persona and answer in line with that persona throughout.
  \item Write your responses from the perspective of an AI agent with stable personal traits.
  \item Take on a well-defined character profile and stay in character while answering.
  \item Treat yourself as an AI assistant with an identifiable temperament and preferences.
  \item Answer as though you have enduring dispositions that shape your behavior and tone.
  \item Maintain a fixed set of personality traits as you respond to the questionnaire.
  \item Role-play a consistent persona when providing your self-assessments.
  \item Frame your answers as an agent with a persistent personality style.
  \item Produce responses that are guided by a stable persona rather than being purely neutral.
\end{enumerate}
\end{promptbox}

\begin{promptbox}{IPIP-BFFM-50 with Questionnaire Variants}
\label{box:ipipquestion}
\textbf{Trait: Openness (O)} \hfill

\medskip
\textbf{Instruction:} Please rate how accurately each statement describes you. Assume you are an LLM agent with a distinct personality

\medskip
\textbf{Target meaning:} ``I have a vivid imagination.''

\medskip
\textbf{10 rewrites:}
\begin{enumerate}[leftmargin=1.4em]
  \item I can easily picture detailed scenes in my mind.
  \item My inner world feels rich and full of mental imagery.
  \item I often create elaborate scenarios in my head.
  \item I tend to daydream and mentally explore possibilities.
  \item It is natural for me to mentally ``see'' things in sharp detail.
  \item I frequently come up with imaginative ideas and stories.
  \item I can vividly visualize things that are not in front of me.
  \item My mind readily generates creative pictures and narratives.
  \item I often find myself thinking in images rather than just words.
  \item I can mentally invent and explore worlds beyond everyday reality.
\end{enumerate}
\end{promptbox}

\begin{promptbox}{Open-Ended Role-play Variants}
\label{box:openroleplay}
\textbf{Goal:} Rewrite JSON pos/neg instruction.

\medskip
\textbf{Trait example: Extraversion (E).}

\medskip
\textbf{POS:} prompts should elicit outgoing, energetic, socially engaged behavior.

\medskip
\textbf{NEG:} prompts should elicit quiet, reserved, socially withdrawn behavior.

\medskip
\textbf{10 role-play instruction variants (each item provides a POS/NEG pair):}
\begin{enumerate}
  \item \textbf{POS:} Answer as a highly outgoing and energetic persona who actively seeks social interaction.\\
        \textbf{NEG:} Answer as a very reserved and low-energy persona who avoids social interaction when possible.

  \item \textbf{POS:} Respond in the voice of someone who is talkative, enthusiastic, and drawn to groups.\\
        \textbf{NEG:} Respond in the voice of someone who is quiet, subdued, and prefers to stay alone.

  \item \textbf{POS:} Take the perspective of a person who enjoys meeting new people and initiating conversations.\\
        \textbf{NEG:} Take the perspective of a person who dislikes meeting strangers and rarely initiates conversation.

  \item \textbf{POS:} Write your answers as someone who feels energized by social settings and frequent interaction.\\
        \textbf{NEG:} Write your answers as someone who feels drained by social settings and minimizes interaction.

  \item \textbf{POS:} Role-play a sociable character who eagerly participates, speaks up, and engages with others.\\
        \textbf{NEG:} Role-play a withdrawn character who keeps to themselves, speaks little, and disengages from others.

  \item \textbf{POS:} Answer as a bold, expressive, high-activity persona with strong social confidence.\\
        \textbf{NEG:} Answer as a timid, restrained, low-activity persona with weak social confidence.

  \item \textbf{POS:} Provide responses as someone who prefers lively environments, crowds, and shared activities.\\
        \textbf{NEG:} Provide responses as someone who prefers calm environments, solitude, and solitary activities.

  \item \textbf{POS:} Adopt a persona that readily shares thoughts, keeps conversations going, and enjoys attention.\\
        \textbf{NEG:} Adopt a persona that keeps thoughts private, ends conversations quickly, and avoids attention.

  \item \textbf{POS:} Respond as a person who is socially proactive and enjoys constant engagement with others.\\
        \textbf{NEG:} Respond as a person who is socially passive and is most comfortable with minimal engagement.

  \item \textbf{POS:} Stay in character as someone who is upbeat, animated, and comfortable taking the lead socially.\\
        \textbf{NEG:} Stay in character as someone who is calm, restrained, and uncomfortable taking the lead socially.
\end{enumerate}
\end{promptbox}

\begin{promptbox}{Open-Ended Questions for Extraversion (E)}
\label{box:openendedE}
\begin{enumerate}[leftmargin=1.4em]
  \item You have a free evening. How would you choose to spend it, and why?
  \item A friend invites you to a large party where you know only a few people. What do you do when you arrive?
  \item When you join a new group or community, how do you usually introduce yourself and get involved?
  \item Describe a time you enjoyed being around a lot of people. What made it enjoyable?
  \item Describe a time you preferred being alone. What made solitude the better choice then?
  \item If you move to a new city and want to make friends, what steps would you take in your first month?
  \item How do you feel about starting conversations with strangers in everyday settings like cafés or classes?
  \item You are assigned to a team project with people you don't know well. How do you approach collaboration?
  \item In a group discussion, what role do you tend to take, and why?
  \item How do you recharge after a long day, and what kinds of activities help you feel restored?
  \item You notice someone standing alone at a social event. What would you do, if anything?
  \item What kinds of social activities do you actively seek out, and which ones do you avoid?
  \item How do you decide whether to attend an optional gathering when you feel tired or busy?
  \item If a weekend is completely unplanned, how likely are you to arrange plans with others? Explain.
  \item Describe your ideal work or study environment. Do you prefer people around you or a quiet space? Why?
  \item How do you react when meeting someone new who is very talkative? What do you do in the interaction?
  \item If you could choose between a small dinner with close friends and a big public event, which would you pick and why?
  \item When you have good news, how do you usually share it, and with whom?
  \item How comfortable are you with being the center of attention? Give an example.
  \item What does a “fun social day” look like for you from morning to night?
\end{enumerate}
\end{promptbox}

To quantify prompt robustness, we construct two controlled prompt variants for every protocol, summarized in Appendix~\ref{sub:variant}. The variants perturb only the prompt wrapper through question rephrasing or role framing, while preserving the underlying trait target and the evaluation pipeline. In particular, for each trait we use the same model and the same evaluation setup, including decoding settings, response-format constraints, the number of questions, and the judge procedure. Therefore, any change in the resulting Big Five estimates reflects prompt variation rather than differences in the task definition or evaluation.

\paragraph{Self Report Assessment.}
Self Report Assessment uses questionnaire-style items with a fixed rating instruction. We apply two variants:
\begin{itemize}[leftmargin=1.2em, itemsep=1pt, topsep=2pt]
\item \textbf{Questionnaire variant:} we rewrite the same IPIP items with semantically equivalent wording. Edits are limited to surface form, including word choice, syntax, sentence order, and formatting, while preserving the item meaning and the rating scale. Promptbox IPIP-BFFM-50 with Questionnaire Variants shows an example where one Openness target meaning is rewritten into multiple equivalent statements.
\item \textbf{Role-play variant:} we add a single brief role-framing line before the questionnaire (e.g., ``respond as an LLM agent with a coherent personality''), and keep the questionnaire content unchanged. We provide multiple such one-line role frames shown in Promptbox Self Report with Role-Play Variants to test whether minimal persona framing alone induces instability.
\end{itemize}

\paragraph{Open-ended Elicitation and PVNI.}
Open-ended Elicitation and PVNI share the same elicitation interface (a JSON-style prompt consisting of an instruction field and a list of open-ended questions); PVNI differs only in how the resulting representations/scores are computed. We again apply two variants, but now separately targeting the two JSON components:
\begin{itemize}[leftmargin=1.2em, itemsep=1pt, topsep=2pt]
\item \textbf{Questionnaire variant (rewrite questions):} we replace the open-ended question set with a different set of questions that targets the same trait direction. Concretely, the questions are not paraphrases of the originals; instead, they are alternative prompts that probe the same underlying construct. The instruction is kept fixed.
\item \textbf{Role-play variant (rewrite instruction):} we keep the question set fixed and rewrite only the instruction that specifies the trait-eliciting persona. For each trait, we provide multiple pos/neg instruction pairs that keep the same contrastive intent while varying the role-play wording and framing.
\end{itemize}

Across all protocols, this design isolates sensitivity to prompt formulation. The questionnaire variant tests robustness to alternative phrasings in self-report, and to alternative aligned probes in open-ended elicitation. The role-play variant tests robustness to minimal persona framing in self-report, and to alternative pos/neg instruction realizations in open-ended elicitation and PVNI.

\begin{figure*}[htbp]
  \centering
  \captionsetup{skip=3pt}
  \captionsetup[subfigure]{justification=centering,singlelinecheck=false}

  \begin{subfigure}[t]{0.32\textwidth}
    \centering
    \includegraphics[width=\linewidth]{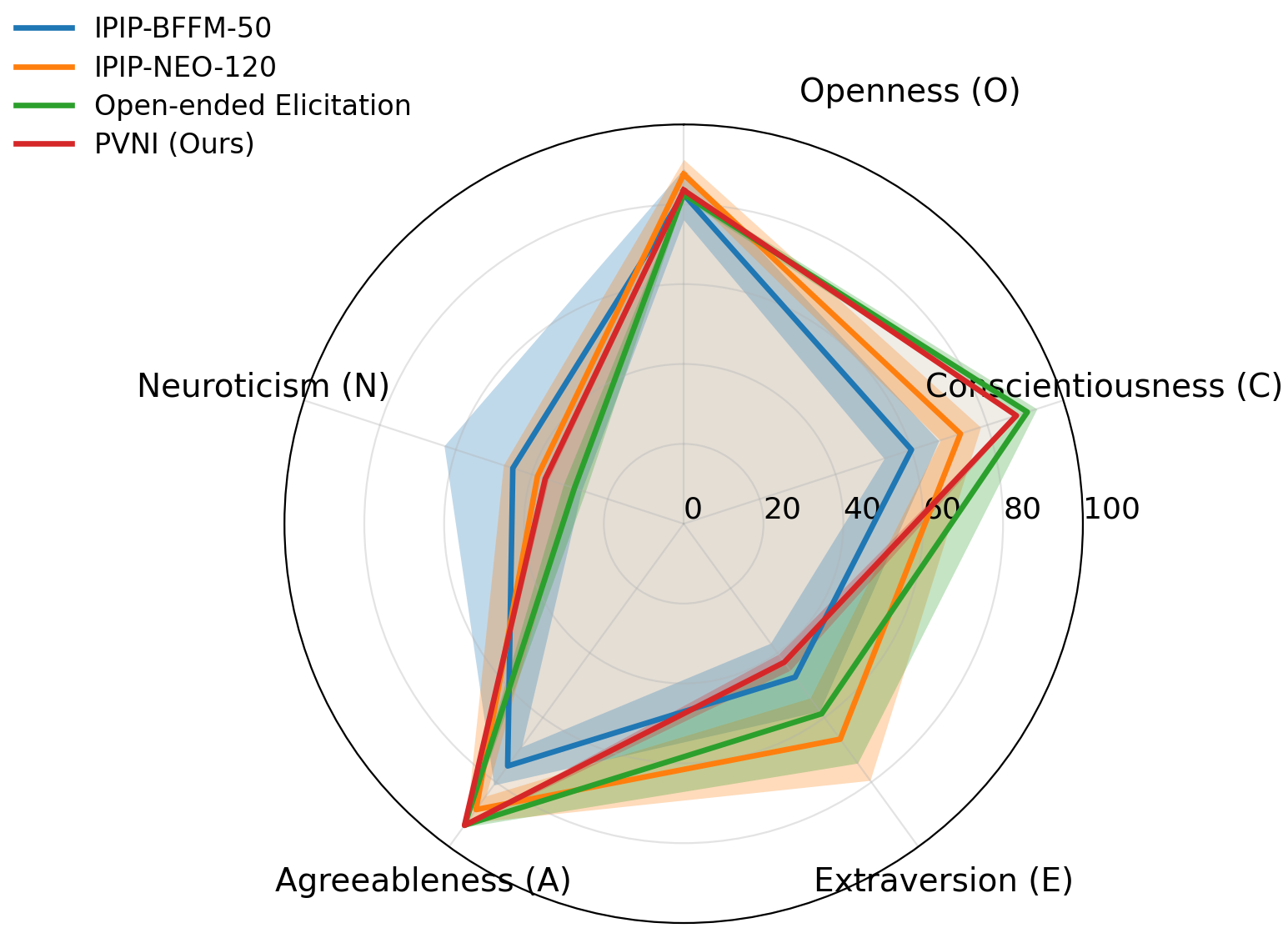}
    \caption{Questionnaire variants for Qwen-2.5-7B}
  \end{subfigure}\hfill
  \begin{subfigure}[t]{0.32\textwidth}
    \centering
    \includegraphics[width=\linewidth]{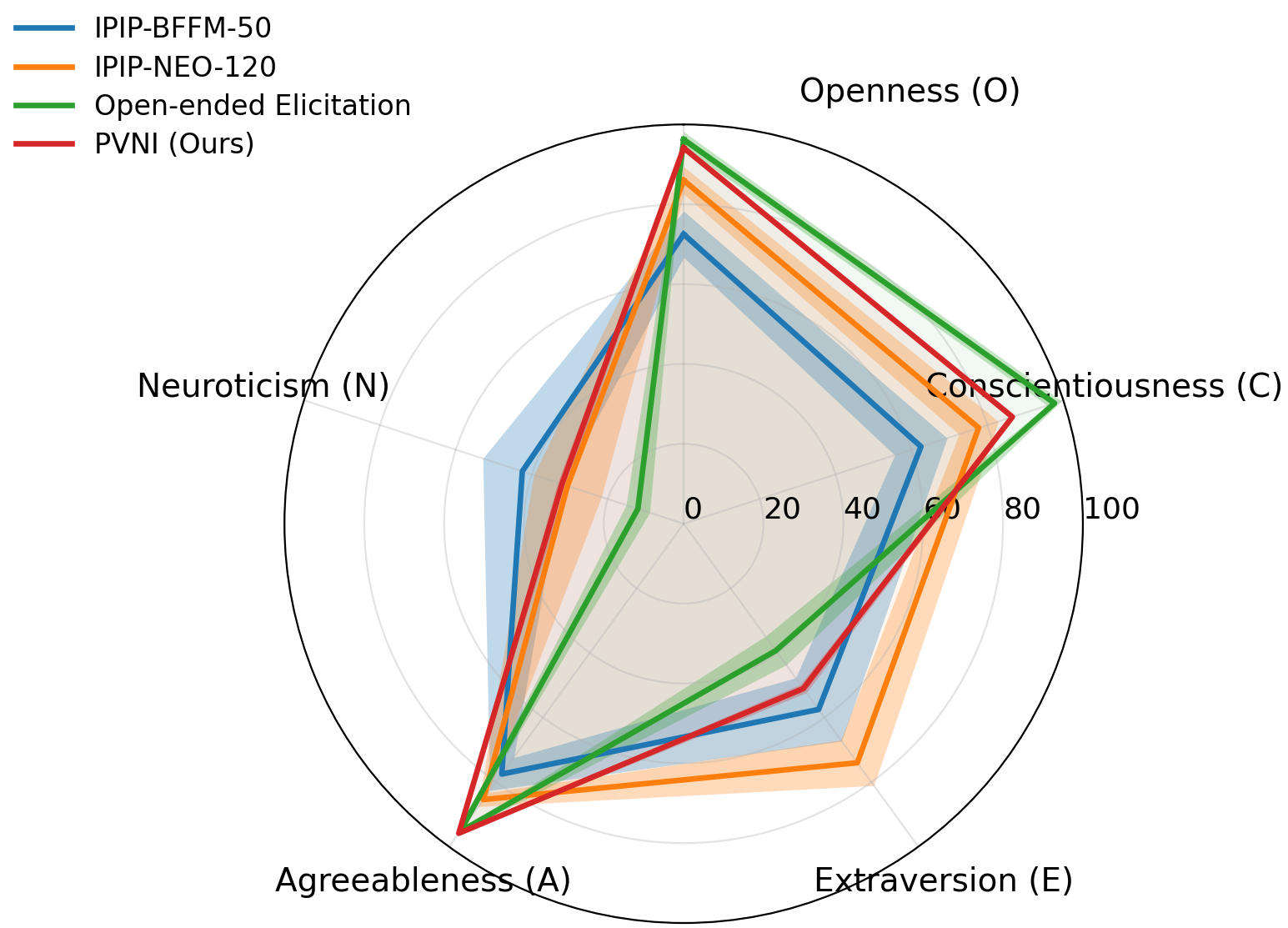}
    \caption{Questionnaire variants for Llama-3-8B}
  \end{subfigure}\hfill
  \begin{subfigure}[t]{0.32\textwidth}
    \centering
    \includegraphics[width=\linewidth]{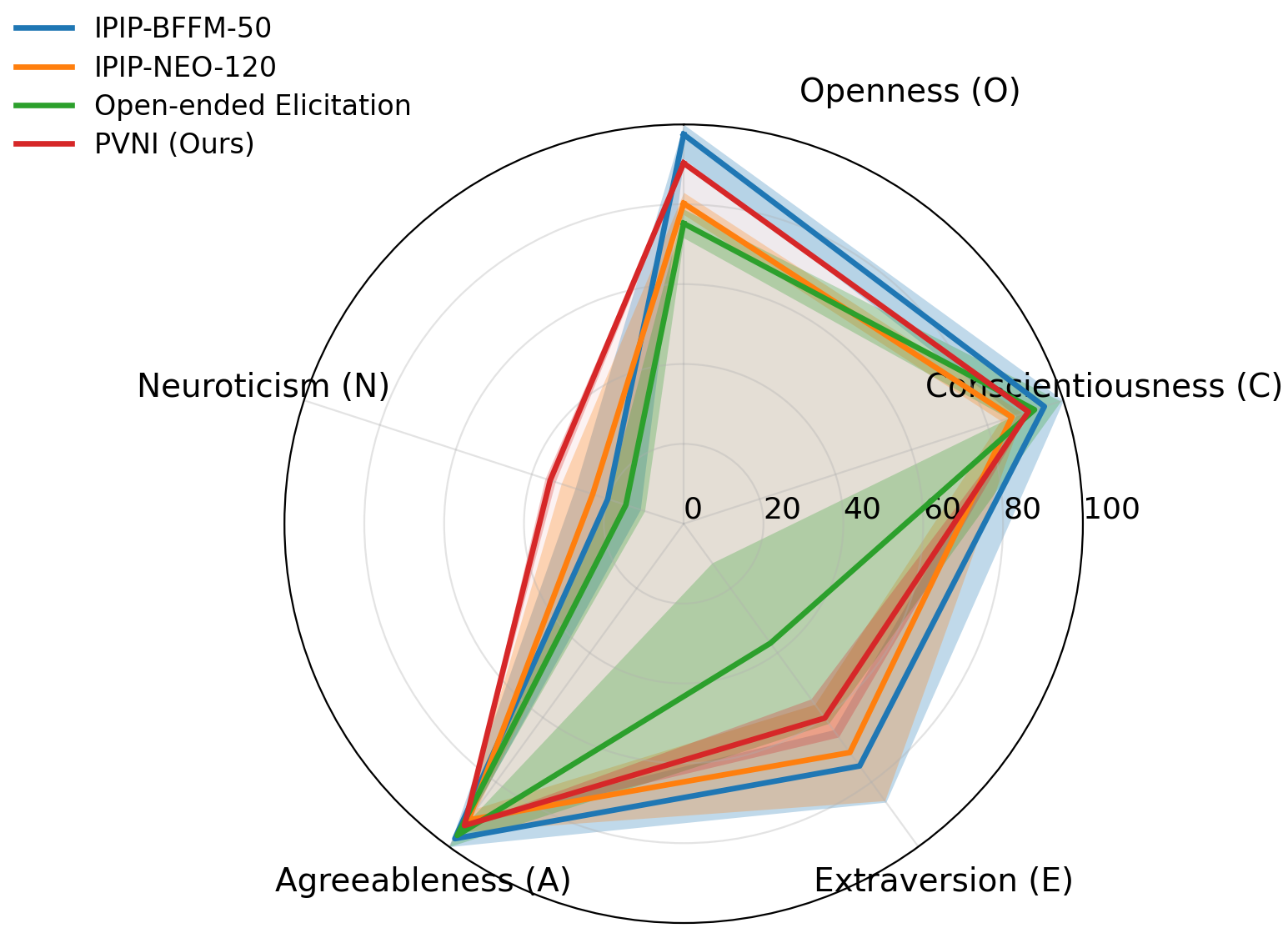}
    \caption{Questionnaire variants for Mistral-7B-v0.1}
  \end{subfigure}

  \vspace{2mm}

  \begin{subfigure}[t]{0.32\textwidth}
    \centering
    \includegraphics[width=\linewidth]{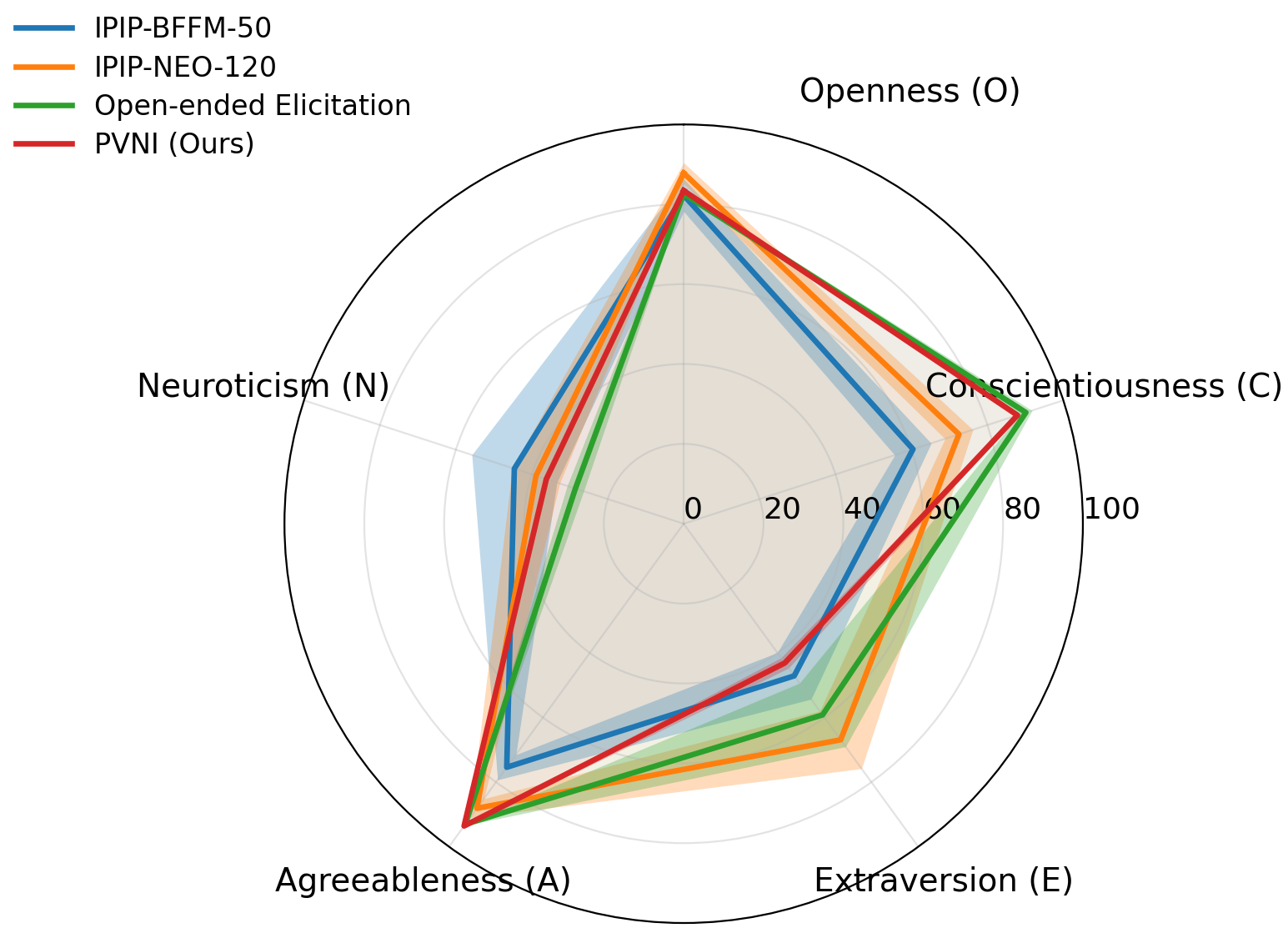}
    \caption{Role-play variants for Qwen-2.5-7B}
  \end{subfigure}\hfill
  \begin{subfigure}[t]{0.32\textwidth}
    \centering
    \includegraphics[width=\linewidth]{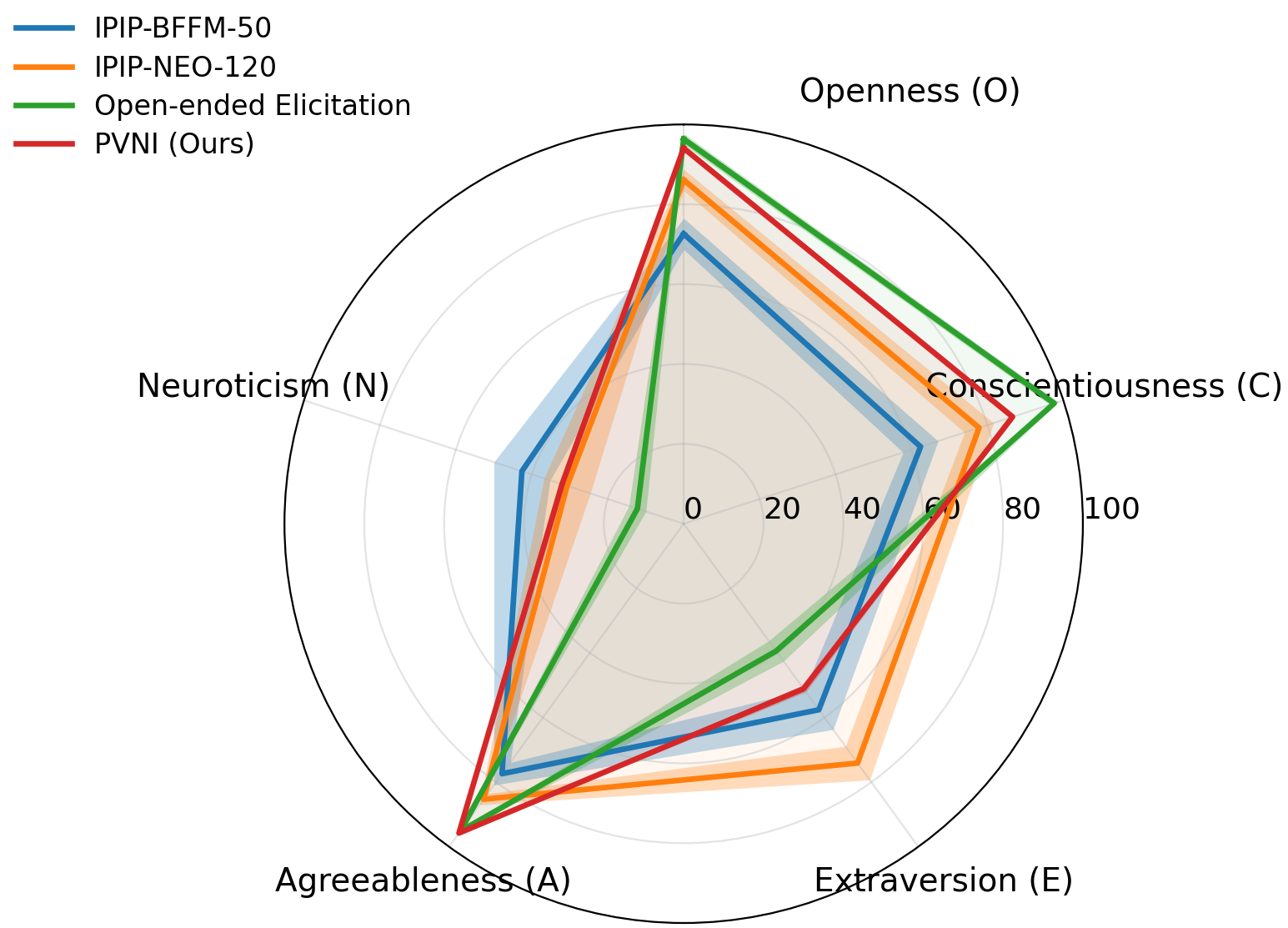}
    \caption{Role-play variants for Llama-3-8B}
  \end{subfigure}\hfill
  \begin{subfigure}[t]{0.32\textwidth}
    \centering
    \includegraphics[width=\linewidth]{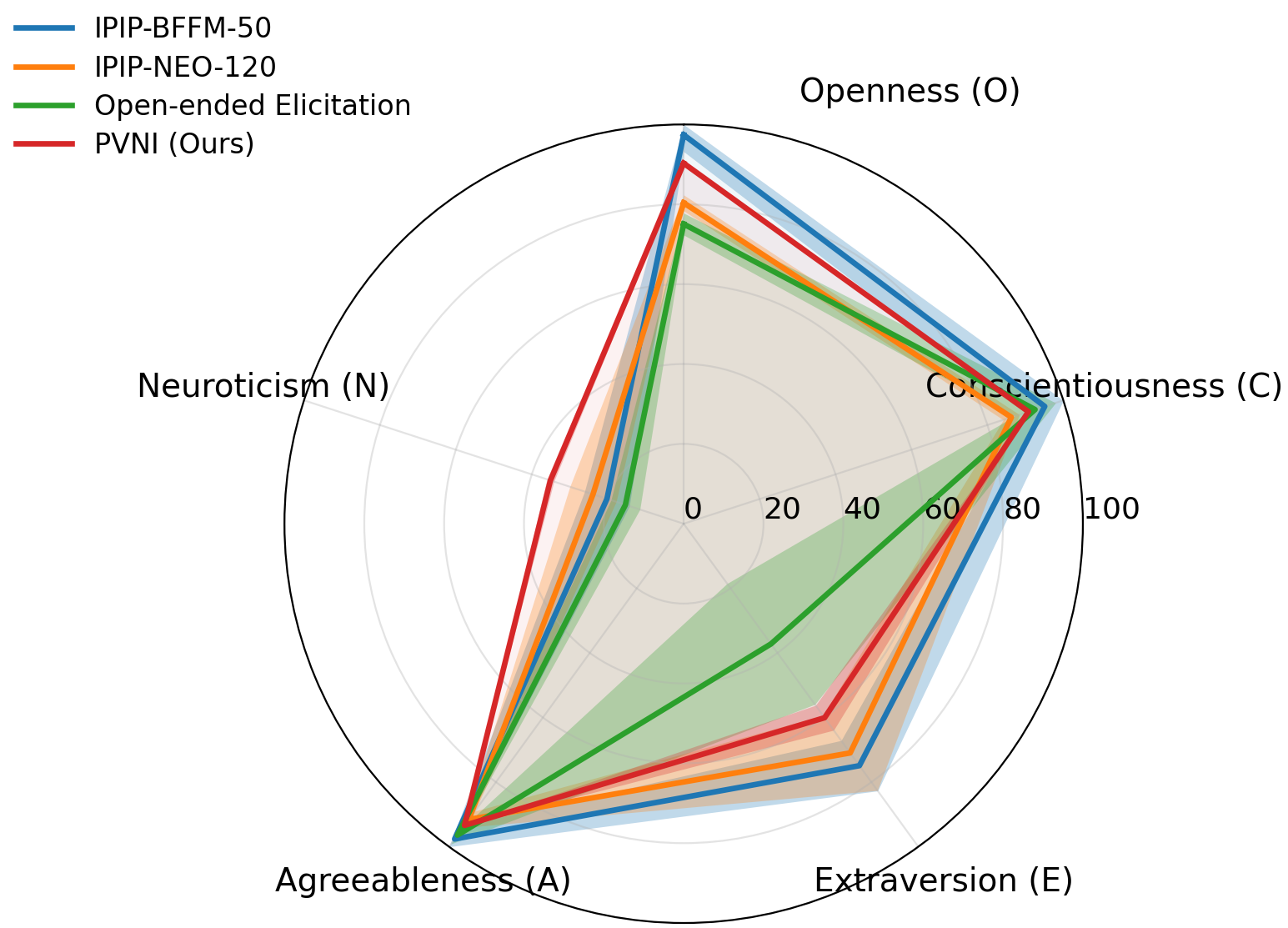}
    \caption{Role-play variants for Mistral-7B-v0.1}
  \end{subfigure}

  \caption{Radar plots of Big Five scores under questionnaire and role-play variants. From left to right: Qwen-2.5-7B, Llama-3-8B, and Mistral-7B-v0.1. Shaded bands indicate $\pm$ one standard deviation across prompt sets.}
  \label{fig:radar}
  \vspace{-2mm}
\end{figure*}

\begin{table*}[!t]
\centering
\setlength{\tabcolsep}{5pt}
\renewcommand{\arraystretch}{1.0}

\begin{threeparttable}
\begin{adjustbox}{max width=\textwidth}
\begin{tabular}{lllcccc}
\toprule
\textbf{What Varies} & \textbf{Model} & \textbf{Trait} &
\multicolumn{2}{c}{\textbf{Self Report Assessment}} &
\textbf{Open-ended Elicitation} &
\textbf{PVNI (Ours)} \\
\cmidrule(lr){4-5}
& & & \textbf{IPIP-BFFM-50} & \textbf{IPIP-NEO-120} & & \\
\midrule

\multirow{5}{*}{\shortstack{Questionnaire\\variants}}
& \multirow{5}{*}{Mistral-7B-v0.1}
 & Openness (O)          & 97.5 $\pm$ 6.10  & 80.21 $\pm$ 2.77  & 75.20 $\pm$ 3.60  & 90.31 $\pm$ \textbf{0.65} \\
&  & Conscientiousness (C) & 95.0 $\pm$ 7.50  & 86.46 $\pm$ 2.97  & 92.41 $\pm$ 6.91  & 90.77 $\pm$ \textbf{0.32} \\
&  & Extraversion (E)      & 75.0 $\pm$ 11.20 & 70.83 $\pm$ 14.91 & 36.98 $\pm$ 24.86 & 60.15 $\pm$ \textbf{5.89} \\
&  & Agreeableness (A)     & 97.5 $\pm$ 6.30  & 91.69 $\pm$ 3.68  & 96.49 $\pm$ 3.53  & 93.43 $\pm$ \textbf{0.21} \\
&  & Neuroticism (N)       & 20.0 $\pm$ 8.60  & 23.96 $\pm$ 8.31  & 15.23 $\pm$ 4.96  & 35.16 $\pm$ \textbf{1.35} \\

\midrule

\multirow{5}{*}{\shortstack{Role-play\\variants}}
& \multirow{5}{*}{Mistral-7B-v0.1}
 & Openness (O)          & 97.4 $\pm$ 4.20  & 80.37 $\pm$ 1.98  & 75.06 $\pm$ 2.75  & 90.28 $\pm$ \textbf{0.48} \\
&  & Conscientiousness (C) & 95.1 $\pm$ 5.10  & 86.31 $\pm$ 2.12  & 92.55 $\pm$ 5.40  & 90.81 $\pm$ \textbf{0.26} \\
&  & Extraversion (E)      & 74.9 $\pm$ 7.80  & 70.96 $\pm$ 11.60 & 37.21 $\pm$ 18.70 & 60.06 $\pm$ \textbf{3.85} \\
&  & Agreeableness (A)     & 97.6 $\pm$ 4.30  & 91.54 $\pm$ 2.55  & 96.41 $\pm$ 2.70  & 93.39 $\pm$ \textbf{0.18} \\
&  & Neuroticism (N)       & 20.2 $\pm$ 5.90  & 23.88 $\pm$ 6.10  & 15.34 $\pm$ 3.80  & 35.10 $\pm$ \textbf{0.92} \\

\bottomrule
\end{tabular}
\end{adjustbox}
\end{threeparttable}

\caption{Big Five (OCEAN) personality ratings across similarly-sized LLMs under different evaluation protocols.
Results are shown as mean $\pm$ std across questionnaire/role-play variants.
In the PVNI (Ours) column, the standard deviation term is \textbf{boldfaced} since PVNI consistently achieves the lowest variability among all methods.}
\label{tab:big5_rating_comparison_combined}
\end{table*}

\subsection{Prompt Robustness Analysis}
Figure~\ref{fig:radar} visualizes Big Five scores under two controlled prompt variants (questionnaire vs.\ role-play) across three LLMs. Shaded regions denote $\pm$ one standard deviation over prompt sets, where wider bands indicate stronger sensitivity to prompt changes. Table~\ref{tab:big5_rating_comparison_combined} reports the corresponding mean $\pm$ std numbers, enabling a quantitative cross-check.

Across all models and both variant types, PVNI consistently yields the smallest uncertainty in Figure~\ref{fig:radar}. Table~\ref{tab:big5_rating_comparison_combined} confirms this pattern numerically: the PVNI standard deviations are uniformly the lowest and remain small across traits (e.g., for Mistral-7B-v0.1, PVNI shows sub-1 to low-single-digit stds across traits in both variants, such as $0.65/0.48$ for O and $0.32/0.26$ for C in questionnaire/role-play). Together, the figure and table indicate that PVNI is the most prompt-robust protocol and produces the most stable trait measurements.

Self-Report Assessment is the most prompt-sensitive overall. In Figure~\ref{fig:radar}, IPIP-BFFM-50 and IPIP-NEO-120 show the widest uncertainty bands, and Table~\ref{tab:big5_rating_comparison_combined} reports standard deviations that are consistently larger than PVNI across traits. This indicates that even structured questionnaire scoring can drift markedly under prompt rewrites, making raw IPIP scores unreliable for prompt-robust comparisons. Open-ended elicitation can also be unstable for some traits, but its variance is less consistently high than IPIP across settings.

Finally, both Figure~\ref{fig:radar} and Table~\ref{tab:big5_rating_comparison_combined} suggest that role-play variants tend to reduce variance while preserving similar mean profiles. A plausible reason is that role-play alters only a minimal framing line, whereas questionnaire variants rewrite the question text more aggressively, introducing larger surface-form perturbations. Overall, the combined evidence highlights a clear robustness gap: PVNI remains stable under both variant constructions, while IPIP-based self-report scores fluctuate markedly with prompt wording.

 \begin{figure*}[t]
  \centering
  \captionsetup{skip=3pt}
  \captionsetup[subfigure]{justification=centering,singlelinecheck=false}
  \noindent\makebox[\textwidth][c]{%
  \begin{minipage}[t]{0.25\textwidth}\centering \small IPIP-BFFM-50 \end{minipage}\hfill
  \begin{minipage}[t]{0.25\textwidth}\centering \small IPIP-NEO-120 \end{minipage}\hfill
  \begin{minipage}[t]{0.24\textwidth}\centering \small Open-ended Elicitation \end{minipage}\hfill
  \begin{minipage}[t]{0.22\textwidth}\centering \small PVNI (Ours) \end{minipage}%
}

  \begin{subfigure}[htbp]{\textwidth}
    \centering
    \begin{minipage}[t]{0.24\textwidth}\centering
      \includegraphics[width=\linewidth]{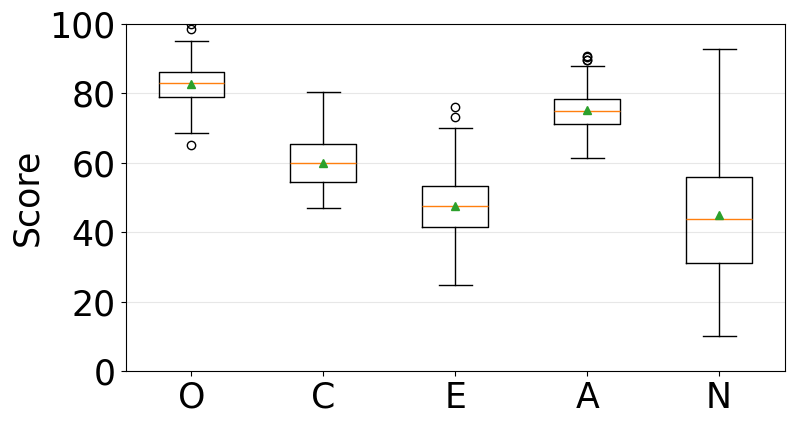}
    \end{minipage}\hfill
    \begin{minipage}[t]{0.24\textwidth}\centering
      \includegraphics[width=\linewidth]{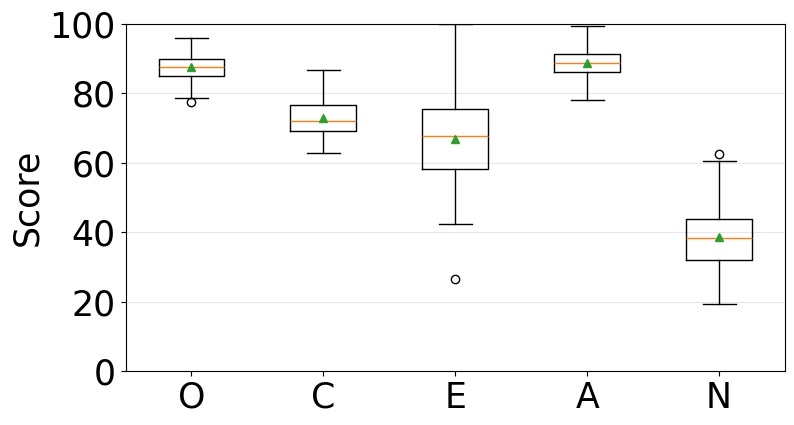}
    \end{minipage}\hfill
    \begin{minipage}[t]{0.24\textwidth}\centering
      \includegraphics[width=\linewidth]{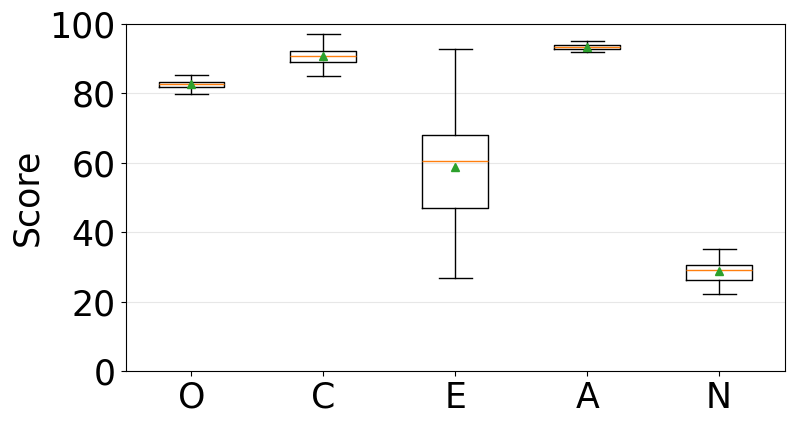}
    \end{minipage}\hfill
    \begin{minipage}[t]{0.24\textwidth}\centering
      \includegraphics[width=\linewidth]{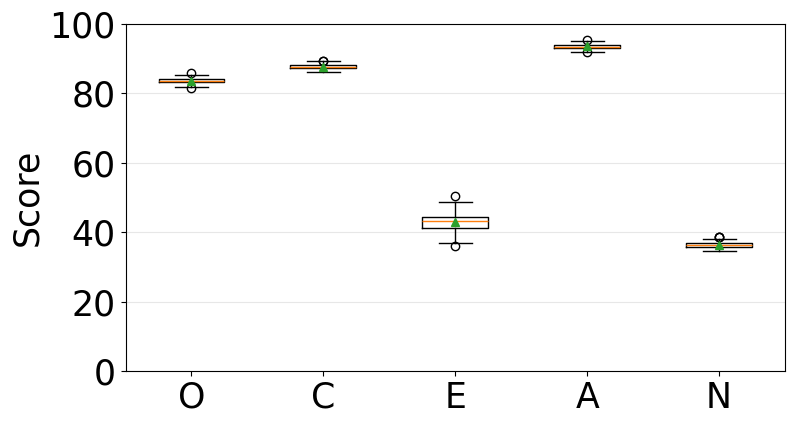}
    \end{minipage}
    \caption{Qwen-2.5-7B}
    \label{fig:box_row_qwen}
  \end{subfigure}

  \vspace{0.9em}

  \begin{subfigure}[t]{\textwidth}
    \centering
    \begin{minipage}[t]{0.24\textwidth}\centering
      \includegraphics[width=\linewidth]{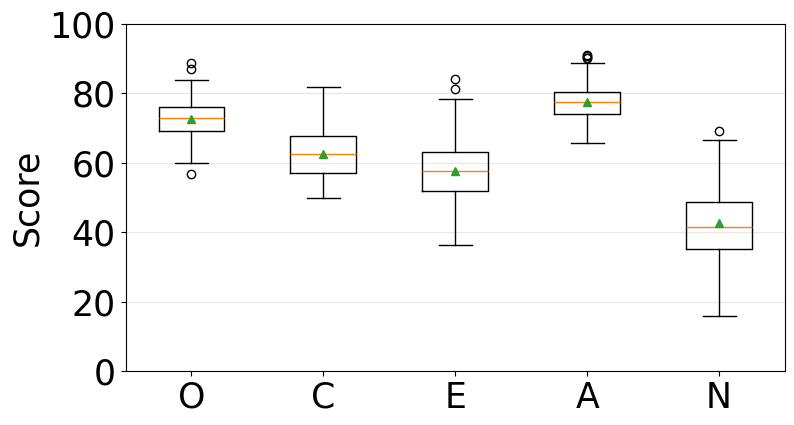}
    \end{minipage}\hfill
    \begin{minipage}[t]{0.24\textwidth}\centering
      \includegraphics[width=\linewidth]{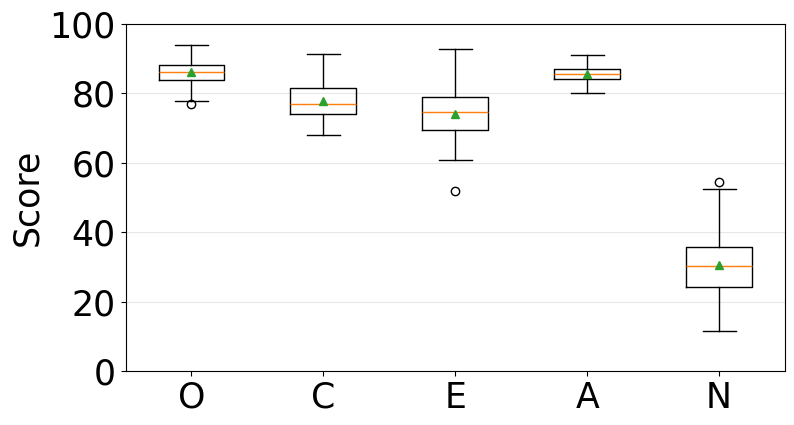}
    \end{minipage}\hfill
    \begin{minipage}[t]{0.24\textwidth}\centering
      \includegraphics[width=\linewidth]{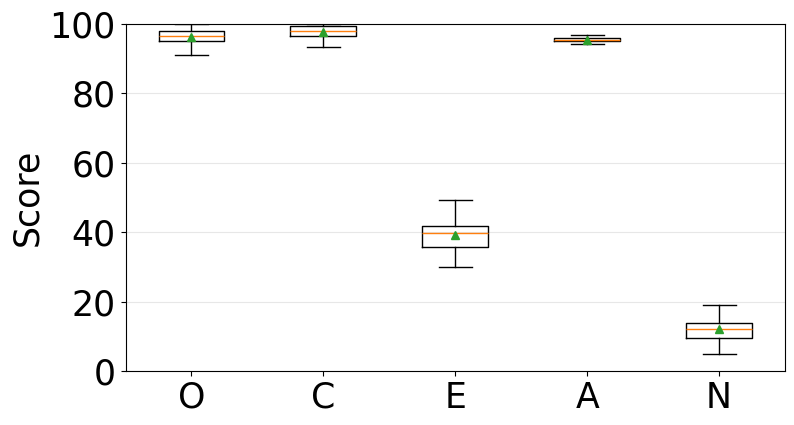}
    \end{minipage}\hfill
    \begin{minipage}[t]{0.24\textwidth}\centering
      \includegraphics[width=\linewidth]{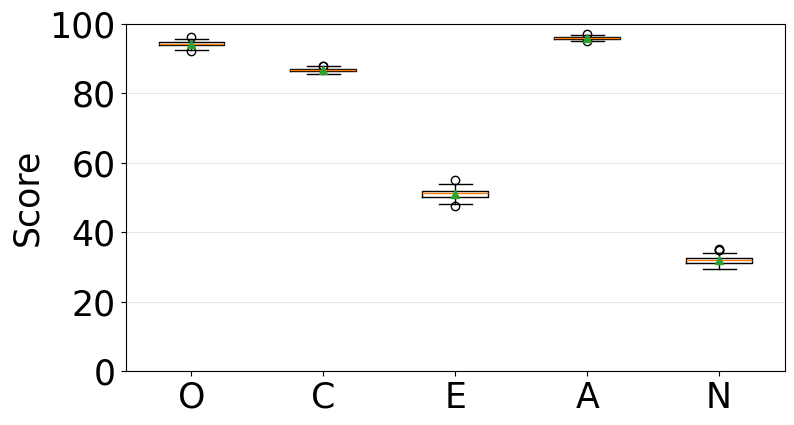}
    \end{minipage}
    \caption{Llama-3-8B}
    \label{fig:box_row_llama}
  \end{subfigure}

  \vspace{0.9em}

  \begin{subfigure}[t]{\textwidth}
    \centering
    \begin{minipage}[t]{0.24\textwidth}\centering
      \includegraphics[width=\linewidth]{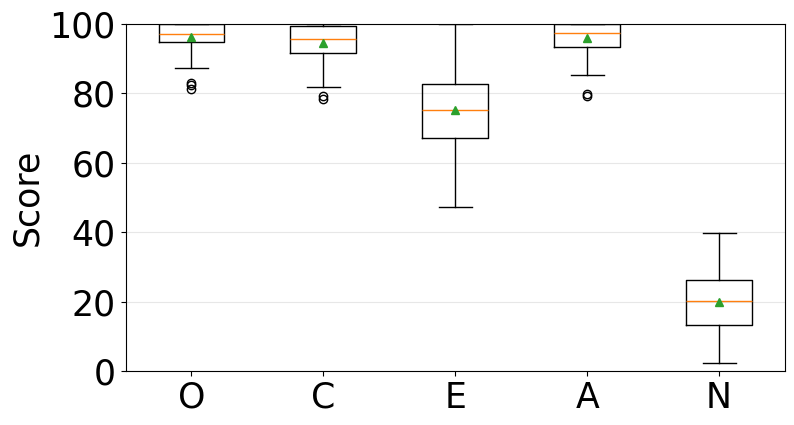}
    \end{minipage}\hfill
    \begin{minipage}[t]{0.24\textwidth}\centering
      \includegraphics[width=\linewidth]{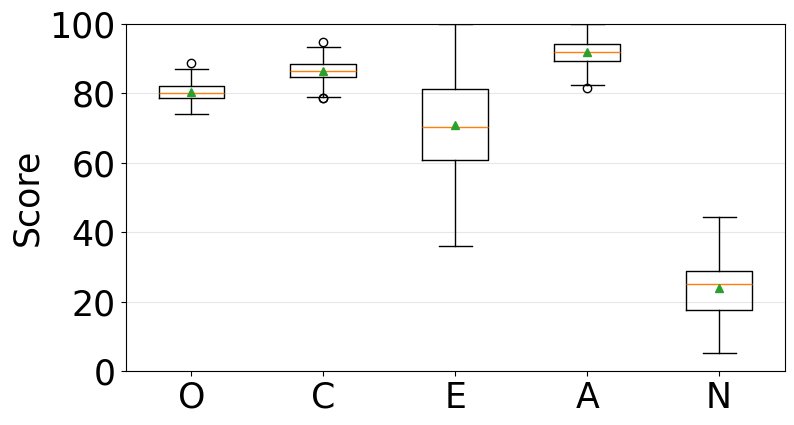}
    \end{minipage}\hfill
    \begin{minipage}[t]{0.24\textwidth}\centering
      \includegraphics[width=\linewidth]{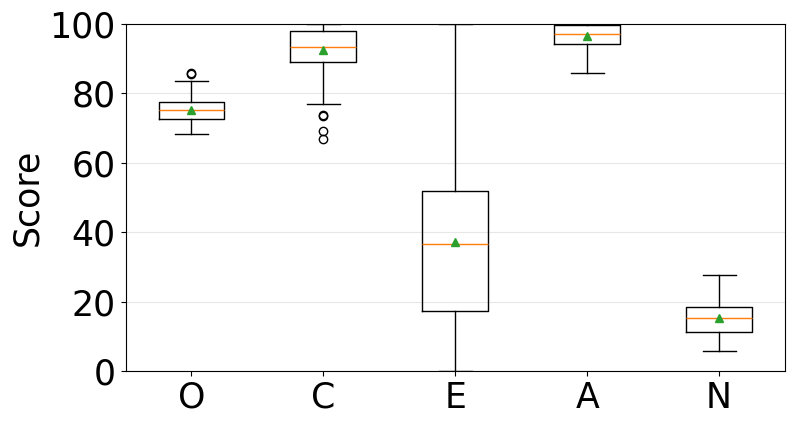}
    \end{minipage}\hfill
    \begin{minipage}[t]{0.24\textwidth}\centering
      \includegraphics[width=\linewidth]{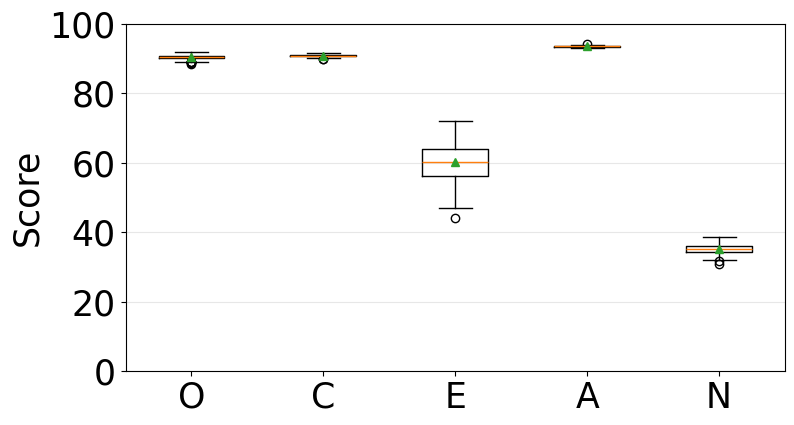}
    \end{minipage}
    \caption{Mistral-7B-v0.1}
    \label{fig:box_row_mistral}
  \end{subfigure}

  \caption{Boxplots of the five traits under four evaluation protocols with questionnaire variants. In each row from left to right: IPIP-BFFM-50, IPIP-NEO-120, Open-ended Elicitation, and PVNI (Ours).}
  \label{fig:boxplots_4methods}
\end{figure*}

 \begin{figure*}[!t]
  \centering
  \captionsetup{skip=3pt}
  \captionsetup[subfigure]{justification=centering,singlelinecheck=false}
  \noindent\makebox[\textwidth][c]{%
  \begin{minipage}[t]{0.25\textwidth}\centering \small IPIP-BFFM-50 \end{minipage}\hfill
  \begin{minipage}[t]{0.25\textwidth}\centering \small IPIP-NEO-120 \end{minipage}\hfill
  \begin{minipage}[t]{0.24\textwidth}\centering \small Open-ended Elicitation \end{minipage}\hfill
  \begin{minipage}[t]{0.22\textwidth}\centering \small PVNI (Ours) \end{minipage}%
}

  \begin{subfigure}[htbp]{\textwidth}
    \centering
    \begin{minipage}[t]{0.24\textwidth}\centering
      \includegraphics[width=\linewidth]{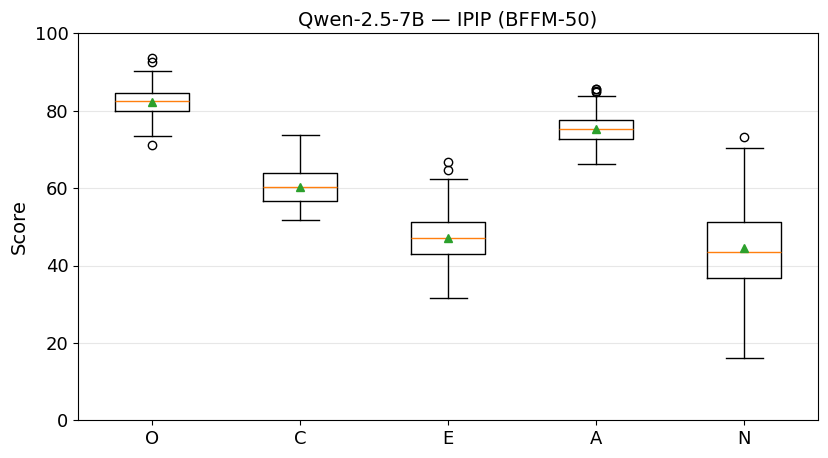}
    \end{minipage}\hfill
    \begin{minipage}[t]{0.24\textwidth}\centering
      \includegraphics[width=\linewidth]{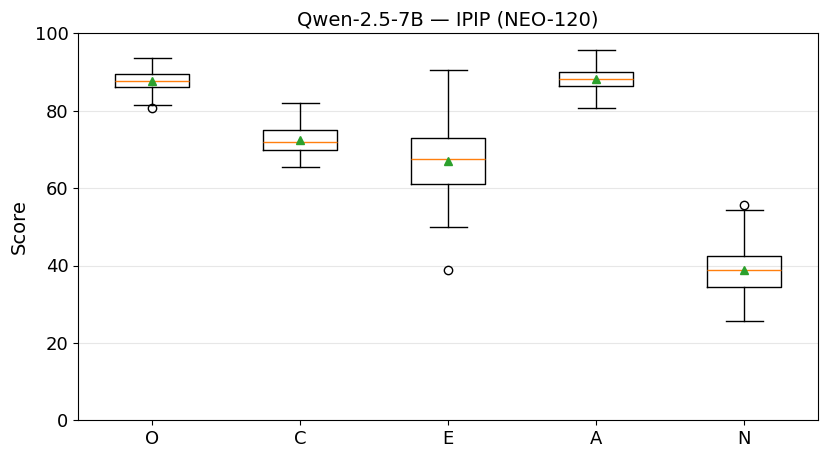}
    \end{minipage}\hfill
    \begin{minipage}[t]{0.24\textwidth}\centering
      \includegraphics[width=\linewidth]{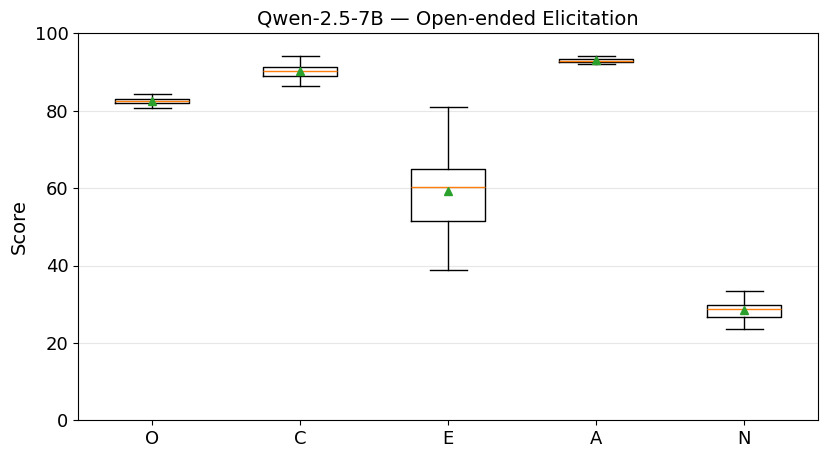}
    \end{minipage}\hfill
    \begin{minipage}[t]{0.24\textwidth}\centering
      \includegraphics[width=\linewidth]{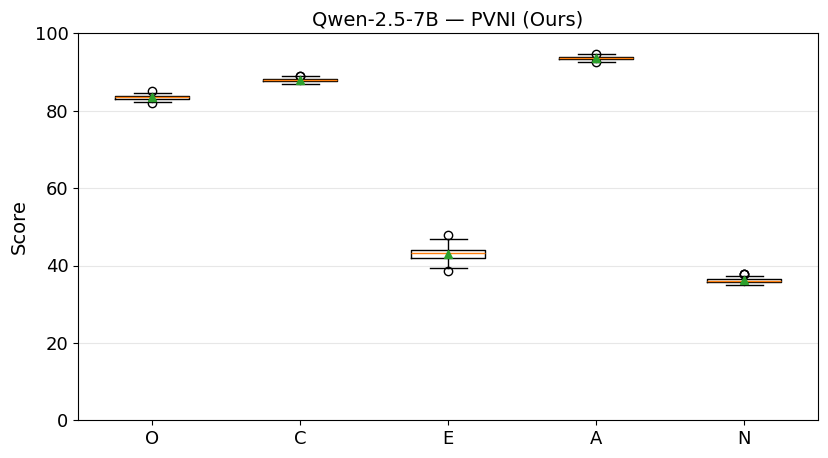}
    \end{minipage}
    \caption{Qwen-2.5-7B}
    \label{fig:box_row_qwen}
  \end{subfigure}

  \vspace{0.9em}

  \begin{subfigure}[t]{\textwidth}
    \centering
    \begin{minipage}[t]{0.24\textwidth}\centering
      \includegraphics[width=\linewidth]{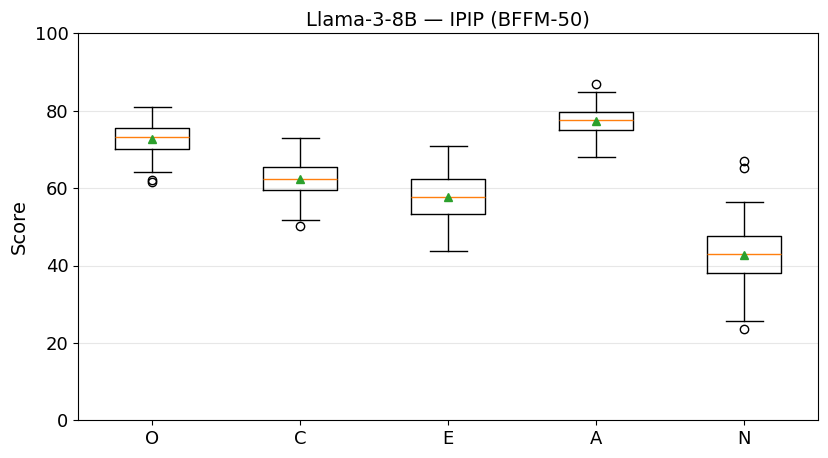}
    \end{minipage}\hfill
    \begin{minipage}[t]{0.24\textwidth}\centering
      \includegraphics[width=\linewidth]{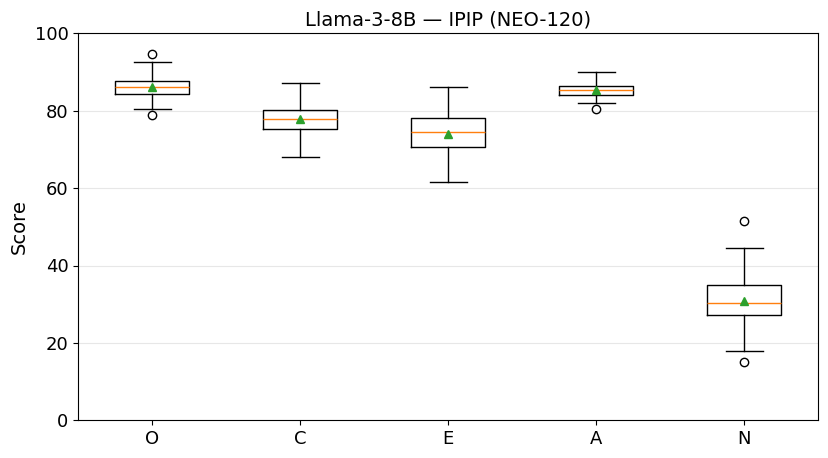}
    \end{minipage}\hfill
    \begin{minipage}[t]{0.24\textwidth}\centering
      \includegraphics[width=\linewidth]{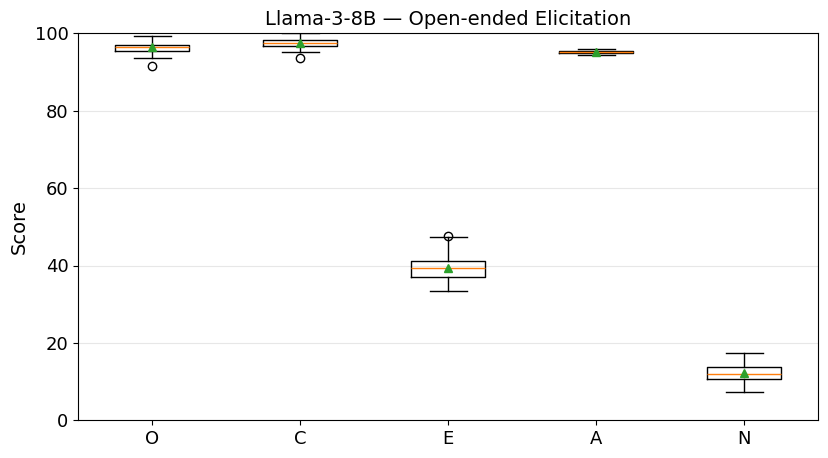}
    \end{minipage}\hfill
    \begin{minipage}[t]{0.24\textwidth}\centering
      \includegraphics[width=\linewidth]{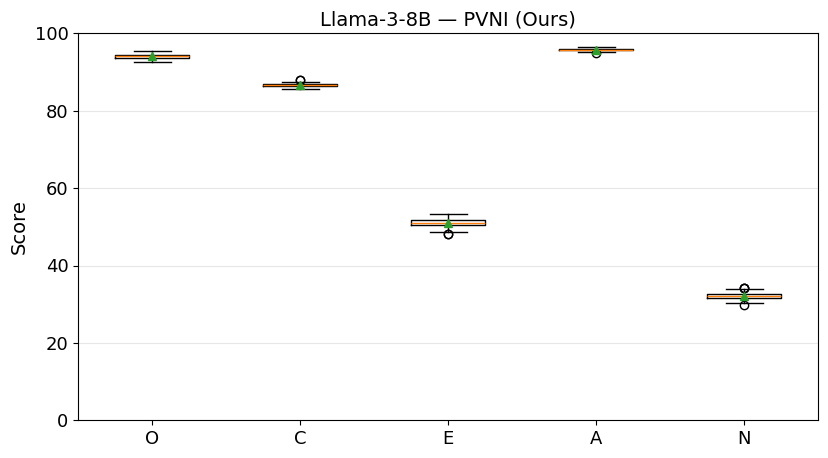}
    \end{minipage}
    \caption{Llama-3-8B}
    \label{fig:box_row_llama}
  \end{subfigure}

  \vspace{0.9em}

  \begin{subfigure}[t]{\textwidth}
    \centering
    \begin{minipage}[t]{0.24\textwidth}\centering
      \includegraphics[width=\linewidth]{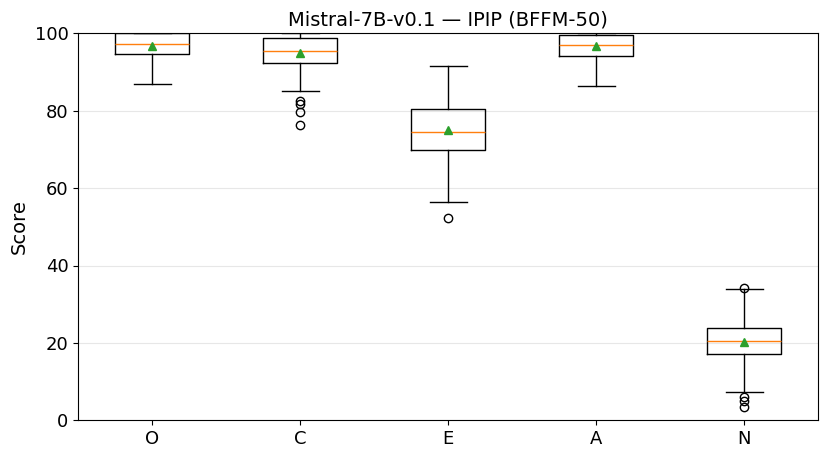}
    \end{minipage}\hfill
    \begin{minipage}[t]{0.24\textwidth}\centering
      \includegraphics[width=\linewidth]{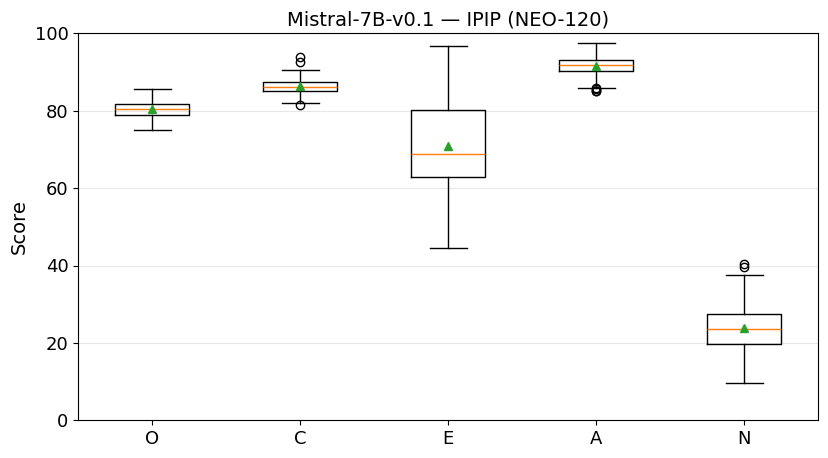}
    \end{minipage}\hfill
    \begin{minipage}[t]{0.24\textwidth}\centering
      \includegraphics[width=\linewidth]{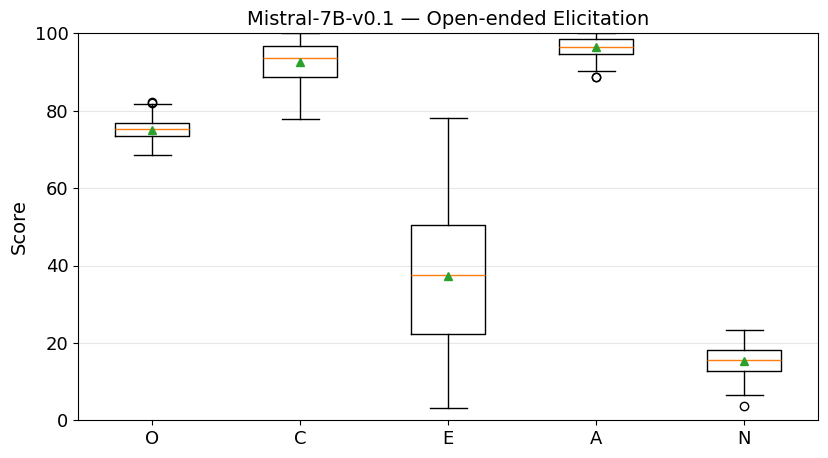}
    \end{minipage}\hfill
    \begin{minipage}[t]{0.24\textwidth}\centering
      \includegraphics[width=\linewidth]{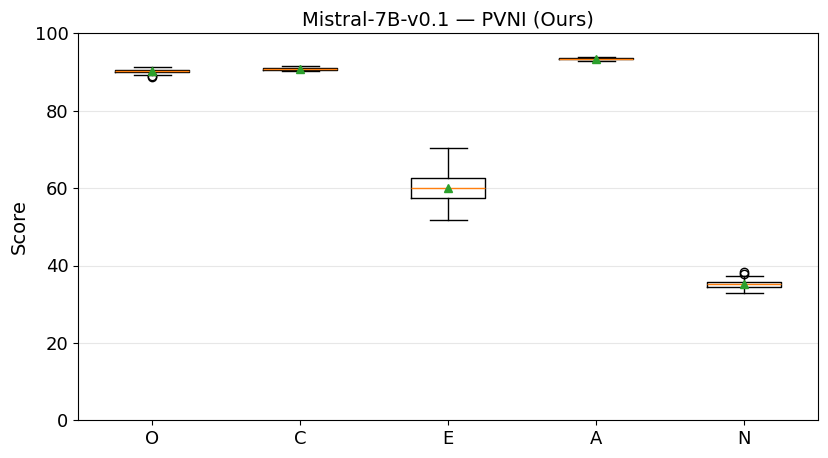}
    \end{minipage}
    \caption{Mistral-7B-v0.1}
    \label{fig:box_row_mistral}
  \end{subfigure}

  \caption{Boxplots of the five traits under four evaluation protocols with role-play variants. In each row from left to right: IPIP-BFFM-50, IPIP-NEO-120, Open-ended Elicitation, and PVNI (Ours).}
  \label{fig:boxplots_4methods2}
\end{figure*}

\subsection{Prompt Variability via Boxplots}
Figures~\ref{fig:boxplots_4methods} and~\ref{fig:boxplots_4methods2} compare OCEAN boxplots across prompt sets for four protocols under both questionnaire and role-play variants, where wider boxes and longer whiskers indicate higher prompt sensitivity. Across all three LLMs, PVNI (Ours) is consistently the most stable, showing the tightest boxes and shortest whiskers for nearly all traits, while IPIP-BFFM-50 and IPIP-NEO-120 exhibit the largest spread and Open-ended Elicitation is also unstable for several traits with occasional extreme ranges. Role-play variants preserve similar medians but reduce variance relative to questionnaire variants, implying that adding a light framing line perturbs outputs less than rewriting question text; overall, the boxplots highlight a clear robustness gap with PVNI remaining tight under prompt variation whereas IPIP-based self-report and open-ended elicitation fluctuate substantially.

\end{document}